\PassOptionsToPackage{unicode}{hyperref}
\PassOptionsToPackage{hyphens}{url}
\documentclass[
]{article}
\usepackage{xcolor}
\usepackage[margin=1in]{geometry}
\usepackage[font=small]{caption}
\usepackage{amsmath,amssymb}
\usepackage{iftex}
\ifPDFTeX
  \usepackage[T1]{fontenc}
  \usepackage[utf8]{inputenc}
  \usepackage{textcomp} 
\else 
  \usepackage{unicode-math} 
  \defaultfontfeatures{Scale=MatchLowercase}
  \defaultfontfeatures[\rmfamily]{Ligatures=TeX,Scale=1}
\fi
\usepackage{lmodern}
\ifPDFTeX\else
\fi
\IfFileExists{upquote.sty}{\usepackage{upquote}}{}
\IfFileExists{microtype.sty}{
  \usepackage[]{microtype}
  \UseMicrotypeSet[protrusion]{basicmath} 
}{}
\makeatletter
\@ifundefined{KOMAClassName}{
  \IfFileExists{parskip.sty}{%
    \usepackage{parskip}
  }{
    \setlength{\parindent}{0pt}
    \setlength{\parskip}{6pt plus 2pt minus 1pt}}
}{
  \KOMAoptions{parskip=half}}
\makeatother
\usepackage{graphicx}
\makeatletter
\newsavebox\pandoc@box
\newcommand*\pandocbounded[1]{
  \sbox\pandoc@box{#1}%
  \Gscale@div\@tempa{\textheight}{\dimexpr\ht\pandoc@box+\dp\pandoc@box\relax}%
  \Gscale@div\@tempb{\linewidth}{\wd\pandoc@box}%
  \ifdim\@tempb\p@<\@tempa\p@\let\@tempa\@tempb\fi
  \ifdim\@tempa\p@<\p@\scalebox{\@tempa}{\usebox\pandoc@box}%
  \else\usebox{\pandoc@box}%
  \fi%
}
\def\fps@figure{htbp}
\makeatother
\usepackage{bookmark}
\IfFileExists{xurl.sty}{\usepackage{xurl}}{} 
\hypersetup{
  hidelinks,
  pdfcreator={LaTeX via pandoc}}

\title{Curriculum unlearning in humans and neural networks}
\author{Jirko Rubruck\textsuperscript{1}*, Alice Zhang\textsuperscript{1}*, Andrew Saxe\textsuperscript{2} and Christopher Summerfield\textsuperscript{1}\\[1ex]
\small \textsuperscript{1} Department of Experimental Psychology, University of Oxford, Oxford, UK\\
\small \textsuperscript{2} Gatsby Unit \& Sainsbury Wellcome Centre, University College London, London, UK\\
\small *These authors contributed equally to this work}
\date{}
\begin{document}
\maketitle

\begin{abstract}
During learning, humans often benefit from curricula that teach simple
rules before exceptions. However, we lack a computational explanation
for this phenomenon. Here, using a canonical behavioural task and
modelling with neural networks, we develop and test a theory for why
simple-to-complex curricula succeed. We show that during learning, both
humans and neural networks show simplicity biases, a tendency to learn
simple rules that are underfit to the data. In neural networks this bias
is accompanied by compression of neural representations.
Simple-to-complex curricula help both types of learners overcome this
compression, developing representations (and policy) with appropriate
dimensionality for the task. We test and confirm a further prediction
from this theory: learners that have been pre-exposed to simple tasks
and with more compressed initial representations benefit more from
curricula. These findings suggest that curricula help with
``unlearning'' of simple rules.
\end{abstract}

\section{Introduction}\label{introduction}

Curriculum learning is the selection and sequencing of training examples
to benefit task acquisition, retention and generalisation. Curricula are
common in natural environments. For example, adult meerkats teach their
offspring to handle dead prey before attempting to catch living
ones\textsuperscript{1}, expert chimpanzees transfer tools to novices to
facilitate termite foraging\textsuperscript{2}, and education programmes
structure information to maximise retention and understanding in human
children. In laboratory experiments, learning of cognitive tasks is
typically boosted by curricula that ``start small'', prioritising simple
elements before complex\textsuperscript{3--5}, typical features before
atypical\textsuperscript{6--9}, or easy examples before
hard\textsuperscript{10--13} (although results for motor skill learning
are more mixed\textsuperscript{14}). Curriculum learning is also of
great practical utility in animal training and
husbandry\textsuperscript{15,16}.

Surprisingly, neural theories struggle to explain why curriculum
learning is successful. In deep networks, which are widely proposed as a
candidate theory for biological learning\textsuperscript{17--19},
curricula do not reliably benefit performance or
generalisation\textsuperscript{20,21}. While some training sequences may
help neural networks compose task primitives\textsuperscript{22}, bridge
the perception-action gap\textsuperscript{23--25}, or direct gradients
towards task-relevant features\textsuperscript{26--29}, these benefits
are typically modest and paradigm-dependent. Accordingly, for supervised
learning tasks that involve assigning labels to stimuli, it is standard
practice to draw independent and identically distributed samples at
random during training\textsuperscript{30}. We currently lack a
generalised account of why curricula work. This holds us back from
designing more effective training regimes for human task learning.

Here, we propose, formalise and test a new theory of curriculum
learning. We focus on a popular approach to curriculum design in which
common (or simple) examples are presented before rare (or complex)
examples (e.g. a language teacher might emphasise generic rules for verb
conjugation before teaching special cases). We study supervised learning
of class labels, a canonical problem in both machine learning and
neuroscience, with widespread translational
application\textsuperscript{31}. Our starting point is that neural
networks trained with gradient descent from small weights (in the
``rich'' regime\textsuperscript{32}) learn dataset modes in order of
importance, such that directions which explain more variance in the data
are learned faster (a `simplicity bias'\textsuperscript{33--38}). As a
result, these networks naturally settle into low-dimensional
representations with a similarity structure that roughly matches that of
the target input-output correlation matrix\textsuperscript{39}.

However, compression comes at a cost. In gradient-based learning,
dimensions that get discarded early are hard to recover later because
they generate near-zero activation and thus weak gradients, preventing
networks from relearning their relevance. Although biological systems do
not learn via gradient descent, they nevertheless show the same striking
difficulty with ``uncompressing'' a dimension that has been learned to
be irrelevant. For example, having initially learned that stimulus
colour is irrelevant to a psychophysical judgment, participants can
perform hundreds of trials without realising that colour is fully
predictive of the correct answer\textsuperscript{40,41}. Brain-computer
interface studies offer an explanation for this phenomenon -- that new
learning is hard outside of the manifold that defines the current neural
activity space\textsuperscript{42}. In other words, once experience has
sculpted neural activity into fixed pathways, new learning is obliged to
respect them, limiting the flexibility of a mature system to form new
representations.

We propose that in both artificial and biological learners, successful
curricula promote learning along dimensions which have been compressed
away and thus lie off manifold\textsuperscript{43}. In other words,
curricula could be said to help more with ``unlearning'' existing
knowledge than with learning from scratch. In what follows, we test this
theory, using a well-studied task that involves discriminating stimuli
(as ``good'' or ``bad'') based on combinations of one, two or three
binary features\textsuperscript{44,45}. Our primary focus is a task
variant in which 75\% of stimuli can be labelled on the basis of a
single feature (``rule'' stimuli) but there exist 25\% stimuli for which
the label is flipped based on a combination of the other features
(``exception'' stimuli). We call this the rule plus exception (R+E)
task, and it corresponds to ``type 5'' in the classification scheme from
Shepard \& Hovland (\textbf{Fig. 1a}).

Past studies of categorisation have repeatedly shown that people learn
rules earlier and more readily than exceptions\textsuperscript{46,47}, a
phenomenon that is captured with hardcoded
models\textsuperscript{48,49}, unsupervised clustering algorithms with
simplicity priors\textsuperscript{50,51}, or connectionist models with
handcrafted basis sets and selective attention\textsuperscript{52}.
Here, we replicate this effect: humans exhibit path dependent learning
and strong initial simplicity biases, with one-dimensional rules applied
early in learning, and more intricate strategies emerging only
later\textsuperscript{53,54}. Although some learners later successfully
learn the exceptions, simplicity biases tend to persist, with
performance on exception stimuli remaining significantly worse than on
rule stimuli even after extensive training. Building on past work with
connectionist models\textsuperscript{55,56}, we show that these findings
all emerge in comparable form during supervised training of deep
networks.

Next, we turn to our main question of how and why curricula affect
learning. Replicating recent findings\textsuperscript{57,58}, we observe
that in both humans and networks, presenting rule stimuli early in
training and exception stimuli later in a ``simple-to-complex''
curriculum leads to improved learning of exceptions. By examining the
representations of our neural network models, we find that the
curriculum helps by facilitating an expansion of the representational
manifold and enabling learners to escape an overly simple,
low-dimensional solution. Correspondingly, networks with more compressed
initial representations benefit more from the \emph{simple-to-complex}
curriculum. Finally, we test if the curriculum helps with uncompression
in human learners by manipulating participants\textquotesingle{} priors
through pretraining on a related task. As in our models, participants
who begin the R+E task with more compressed, low-dimensional priors
benefit more from curricula. Together, these findings support the claim
that \emph{simple-to-complex} curricula work by facilitating the
uncompression of dimensions that prior learning has rendered irrelevant
-- a phenomenon we dub `curriculum unlearning'.

\section{Results}\label{results}

\subsection{Humans and networks demonstrate simplicity
biases}\label{humans-and-networks-demonstrate-simplicity-biases}

\begin{figure}[htbp]
\centering
\includegraphics[width=0.8\linewidth]{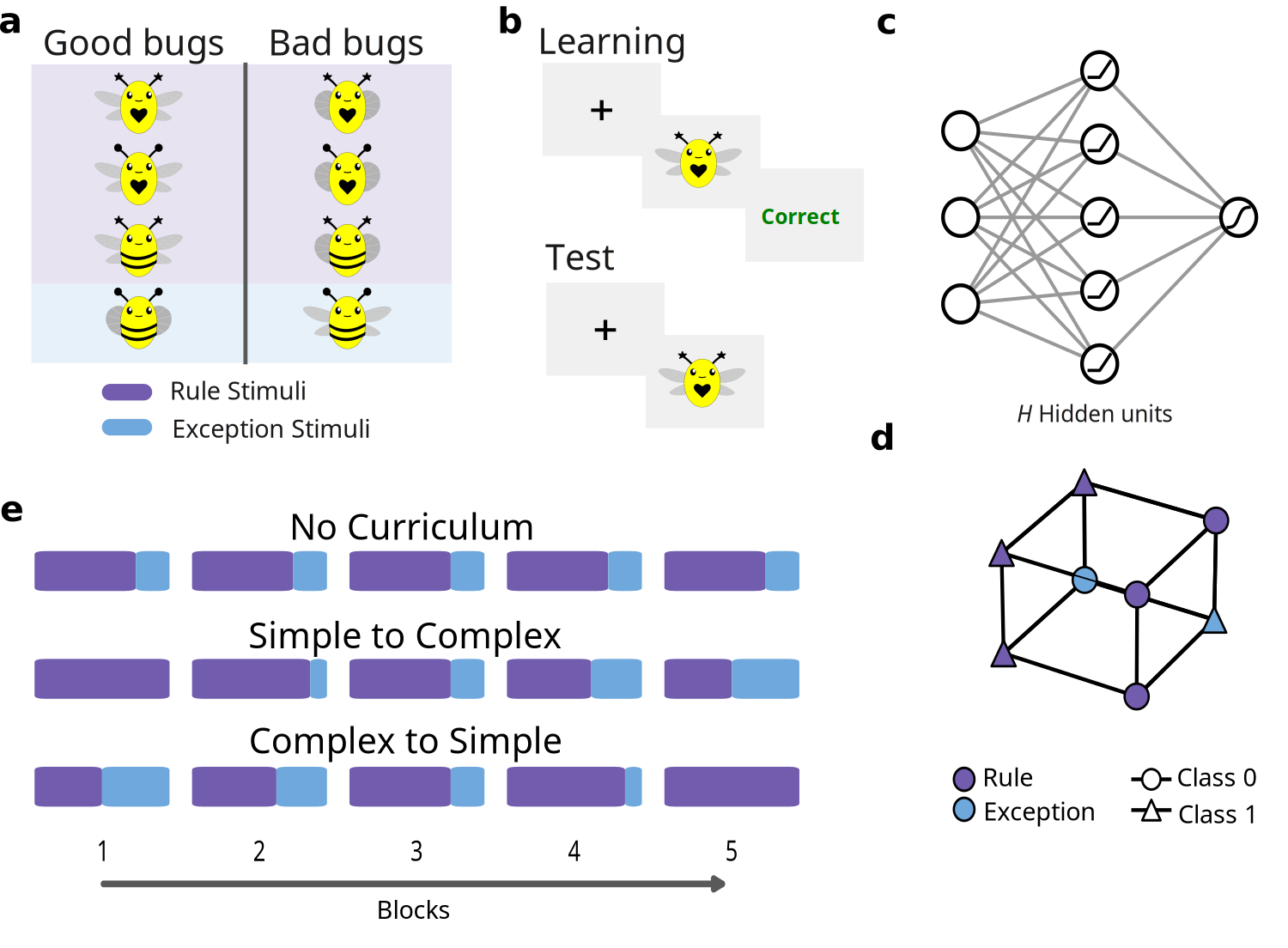}
\caption{\textbf{a}, Participants learned to classify bug stimuli as
either ``good'' or ``bad'' in a rule plus exception (R+E) task. Most
bugs can be correctly classified based on a single feature (here wings)
but there are two exceptions to this rule. \textbf{b}, Participants
learned the task in a trial-by-trial manner with fully supervised
feedback. After the training phase, they were assessed in a test phase
without feedback. \textbf{c}, We modelled the task using a two-layer
neural network with stimuli (\textbf{d}) which similarly varied along
three binary dimensions. The eight stimuli can be depicted as the
corners of a cube in a three-dimensional binary feature space. Colour
denotes rule (purple) versus exception (blue) stimuli and shape denotes
the category label (circle = Class 0, triangle = Class 1). The network
(bottom) contained H ReLU hidden units and a sigmoid output unit and was
trained using trial-wise gradient descent, we trained one network per
participant on that participant\textquotesingle s own sequence of
trials, so that each network receives identical stimuli in identical
order to human participants \textbf{e}, Training consisted of five
learning blocks of 48 trials each in which rule and exception stimuli
were randomly interleaved. In the ``no curriculum'' condition, stimuli
were sampled pseudorandomly on each block (36 rule stimuli and 12
exceptions). In the \emph{simple-to-complex} condition, the fraction of
exception stimuli in each block increased over the course of five blocks
from 0 to 0.5. In the ``complex-to-simple'' curriculum, the fraction of
exception stimuli decreased over learning blocks. All curriculum
conditions are matched for the overall number of exposures to each
stimulus.}
\end{figure}

In Exp. 1, n = 207 human participants learned to classify bug stimuli
defined by three binary features (body pattern, antennae type, and wing
shape; \textbf{Fig. 1a}). Six of the eight stimuli could be classified
by a rule based on a single feature (rule stimuli), whereas the
remaining two violated this rule (exception stimuli). Participants were
randomly assigned to one of three curriculum conditions that varied the
order in which rule and exception stimuli were presented (\textbf{Fig.
1e}). After learning via trial and error over five blocks of 48 trials
(\textbf{Fig. 1b}), participants completed a test phase in which they
classified each stimulus twice without feedback.

\begin{figure}[htbp]
\centering
\includegraphics[width=\linewidth]{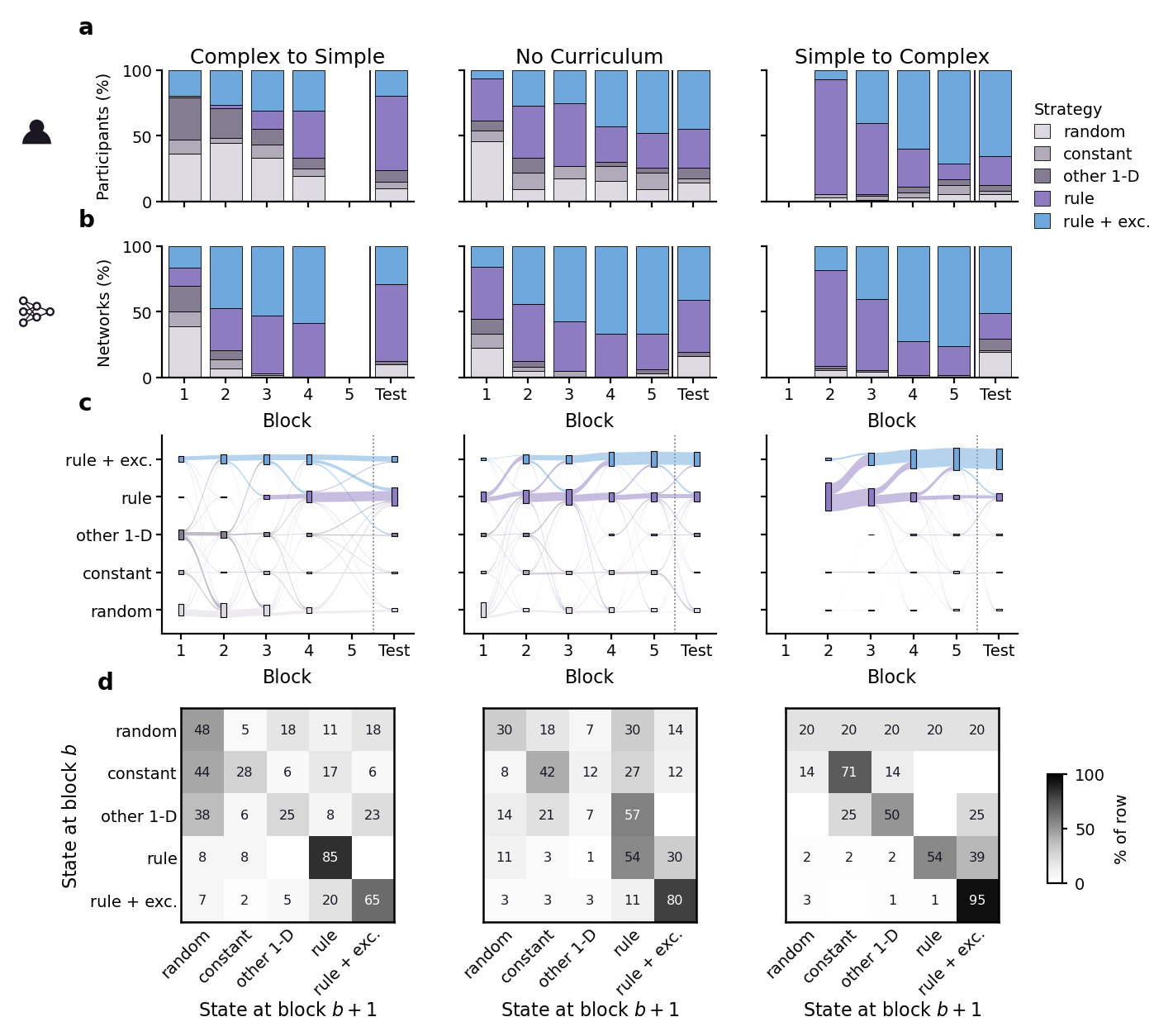}
\caption{\textbf{a,} Proportion of participants (n = 207) classified
into each of five strategies at the end of each learning block and at
test, separately by curriculum. Strategies were assigned by fitting nine
deterministic policies plus a lapse rate and selecting by BIC against an
unpenalised random-responding null. \textbf{b,} The same classification
applied to networks, one per participant, trained on that
participant\textquotesingle s exact trial sequence. Icons mark which
rows show people and which show networks. \textbf{c,} Block-to-block
flow of human strategies, with ribbon width proportional to the number
of participants making each transition; the dotted vertical line marks
the test phase. \textbf{d,} Transition matrices summarising the
transitions in c, across training blocks giving the percentage of
participants in each state at block b who occupied each state at block
b+1. Rows sum to 100\%. Pooled across curricula, the odds of moving from
rule to rule + exceptions were 2.23 times higher under the
\emph{simple-to-complex} curriculum than under the other two (95\% CI
{[}1.18, 4.24{]}, \emph{p =} .014; Mantel--Haenszel stratified by block,
participant-clustered bootstrap {[}1.21, 4.51{]}).}
\end{figure}

\subsubsection{Humans show simplicity
biases}\label{humans-show-simplicity-biases}

We first used a model-fitting approach to study the strategies used
during learning. For every participant and block we asked which of ten
candidate policies best explained responses including the correct policy
(``rule + exc''), adherence to the rule only (``rule''), other incorrect
1D rules (``other 1D'', 5 variants), constant responding with a single
key (``constant'', 2 variants), and random responding (``random''). Each
policy was fitted with a maximum-likelihood lapse rate, so that
consistent but imperfect behaviour is still attributed to the policy it
follows, and policies were compared by BIC against the random-responding
null, which was left unpenalised. \textbf{Fig. 2a} (top row) shows the
percentage of humans best classified under each strategy for each block
during training and the rightmost bar shows classifications at test for
each of the three curricula. Note that strategies can only be
distinguished where both rule and exception stimuli occur, so the first
and last bar of the \emph{simple-to-complex} and
\emph{complex-to-simple} curriculum respectively are missing.

Under \emph{no curriculum} (\textbf{Fig. 2a}, middle), early blocks are
dominated by the rule-only strategy, which peaks at 47.6\% of
participants by block 3 while only 25.4\% are following rule plus
exceptions. Of the 46 participants using one of the two rule-based
strategies, 30 used the rule alone (exact binomial \emph{p} = .054).
This is a simplicity bias -- participants initially learned the simple
rule that explains a large amount of dataset variance, before
transitioning to a more complex strategy\textsuperscript{46,47}.
However, this bias is substantially mitigated by the
\emph{simple-to-complex} curriculum. Participants there also begin
strongly rule-biased, since only rule stimuli are present in the first
block. We find that 87.5\% follow a rule-only strategy in block 2.
However, the gradual introduction of exceptions reverses this pattern.
The proportion following a rule + exc strategy dramatically increases
from 6.9\% to 70.8\% across blocks 2 to 5, and reaches 65.3\% at test
time compared to 44.4\% in the \emph{no curriculum} condition (+20.8
percentage points {[}4.0, 36.1{]}, OR = 2.35, Fisher\textquotesingle s
exact \emph{p} = .024), while only 19.4\% reached it under
\emph{complex-to-simple} (\ensuremath{-}25.0 points {[}\ensuremath{-}39.3, \ensuremath{-}9.3{]}, OR = 0.30,
\emph{p} = .005).

Inspection of \textbf{Fig. 2} suggests a twofold benefit. Firstly,
relatively fewer participants stick with the rule strategy, failing to
learn the exceptions (at test 22.2\% under \emph{simple-to-complex}
against 30.2\% under \emph{no curriculum} and 56.9\% under
\emph{complex-to-simple};\(\chi^{2}(2)\) = 20.36, \emph{p} \textless{}
.001, \emph{V} = 0.31). Secondly, it seems as if fewer participants
pursue idiosyncratic, unrelated strategies or random responding (12.5\%
of participants were classified as random, constant, or following some
other one-dimensional rule, against 25.4\% and 23.6\% respectively).
However, this difference is not significant (\(\chi^{2}(2)\) = 4.21,
\emph{p} = .12). One might ask whether graduating the stimulus types
(e.g. separating rule and exception training in time) was sufficient to
provoke these effects. It seems not, because in the
\emph{complex-to-simple} curriculum relatively fewer participants
acquire the optimal (rule + exc) strategy across the course of training.
Specifically, under \emph{complex-to-simple} we find a drift to the
rule-only strategy as exceptions are withdrawn, 1.4\% of participants
follow rule-only in block 1, rising to 36.1\% by block 4.

Another way to show the simplicity bias and its mitigation by a
\emph{simple-to-complex} curriculum is to count the blockwise
transitions between strategies, which we visualise in \textbf{Fig. 2c}
and tabulate in \textbf{Fig. 2d}. First, under no curriculum the bias
has a temporal dynamic such that participants pass through the rule-only
state before acquiring the optimal strategy, with entry into rule +
exceptions over four times more likely from the rule state than from
other states (30.4\% against 11.5\%; OR = 4.15 {[}1.80, 9.55{]}, \emph{p
\textless{}} .001). Under \emph{simple-to-complex} curriculum
participants enter the optimal strategy more readily (39.0\% of
rule-only blocks were followed by a rule-plus-exception block, against
30.4\% under \emph{no curriculum} and 0 of 13 participants under
\emph{complex-to-simple}). Pooling across block pairs and stratifying by
block, the odds of this transition were 2.23 times higher under the
\emph{simple-to-complex} curriculum than under the other two (95\% CI
{[}1.18, 4.24{]}, \emph{p} = .014; Mantel--Haenszel stratified by block,
participant-clustered bootstrap {[}1.21, 4.51{]}). Second, the strategy
is significantly more stable under \emph{simple-to-complex}, such that
94.8\% of participants using the rule+exc strategy state remained there
at the next block, against 79.7\% under \emph{no curriculum} and only
65.5\% under \emph{complex-to-simple} (OR = 3.63 {[}1.12, 11.74{]},
\emph{p} = .025).

\subsubsection{\texorpdfstring{Simple-to-complex curricula facilitate
learning
}{Simple-to-complex curricula facilitate learning }}\label{simple-to-complex-curricula-facilitate-learning}

The \emph{simple-to-complex} curriculum encourages more participants to
behave in a way that is consistent with the rule+exc strategy, and
should thus benefit learning. To verify this, we plotted human accuracy
across training (\textbf{Fig. 3a}) and test (\textbf{Fig. 3b}) under
each of the three curricula, separately for rule and exception stimuli.
During training, differences in performance between conditions may be
driven by patterns of stimulation under distinct curricula, so we focus
on the \emph{no curriculum} condition. We observe three salient effects.
Firstly, rule stimuli are classified more accurately than exceptions
(Log-Odds = -0.48 {[}-0.52 -- -0.44{]}, \emph{z} = -24.04, \emph{p}
\textless{} .001). Secondly, across the course of training, participants
improved more on rule than exception stimuli, although this effect was
marginal (Log-Odds = -0.04 {[}-0.08 -- 0.00{]}, \emph{z} = -1.97,
\emph{p} = .049; \textbf{Fig. 3a}). Thirdly, performance on the
exceptions dips below chance early in training (\textbf{Fig. 3a}, first
block: M = 0.44, 95\% CI {[}0.39 0.48{]}, \emph{t(}62) = -2.79, \emph{p}
= 0.007, \emph{d} = -0.35, which is expected if participants are
displaying a simplicity bias that misclassifies exception stimuli using
the 1-D rule.

\begin{figure}[htbp]
\centering
\includegraphics[width=\linewidth]{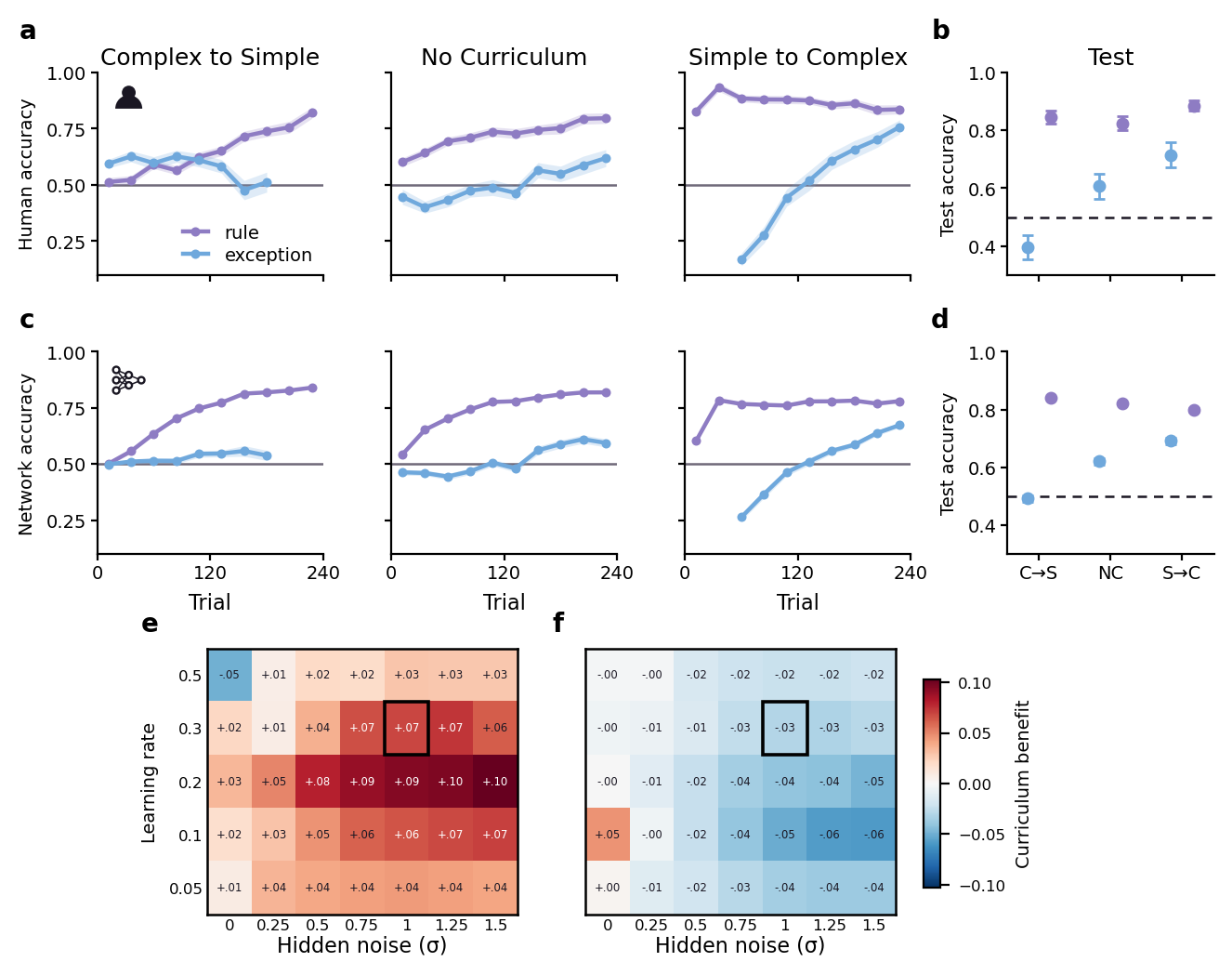}
\caption{\textbf{Simple-to-complex curricula boost exception learning
in humans and neural networks.} \textbf{a,} Human accuracy over
training, binned into ten 24-trial bins, separately for rule (purple)
and exception (blue) stimuli and for each curriculum. Shading gives \ensuremath{\pm}1
SEM across participants. \textbf{b,} Human test accuracy, averaged over
two presentations of each of the eight stimuli. \textbf{c}, \textbf{d}
The same for the networks, measured identically, per-trial accuracy on
the presented stimulus during training, and noise-marginalised accuracy
at test. Networks inherit each participant\textquotesingle s trial
sequence, so the two rows are matched trial for trial and differ only in
the learner. Error bars give \ensuremath{\pm}1 SEM. \textbf{e, f} Hyperparameter sweep
for the networks. Colour gives the curriculum benefit ---
simple-to-complex minus no curriculum --- in end-of-training accuracy on
exception stimuli (\textbf{e}) and rule stimuli (\textbf{f}). The black
box marks the configuration used throughout the paper. The exception
benefit is positive in 34 of 35 configurations, so the effect does not
depend on the exact hyperparameters selected.}
\end{figure}

The test phase is matched across curriculum conditions, allowing us a
clean comparison of performance (\textbf{Fig. 3b}). Participants who
learned under the \emph{simple-to-complex} curriculum performed better
than those who received \emph{no curriculum} (Log-Odds = 0.78 {[}0.20,
1.36{]}, \emph{z} = 2.64, \emph{p} = .008). This coefficient is the
log-odds difference at the average of the two stimulus types, because
stimulus type was sum-coded. To break this down, we computed estimated
marginal means contrasts within each stimulus type. The
simple-to-complex advantage was present for both exception stimuli
(Log-Odds = 0.93 {[}0.02, 1.83{]}, \emph{z} = 2.01, \emph{p} = .045) and
rule-consistent stimuli (Log-Odds = 0.63 {[}0.02, 1.24{]}, \emph{z} =
2.02, \emph{p} = .044). The exception benefit is consistent with prior
work on the rule-plus-exception task, in which delaying exceptions until
after rule-consistent items improved exception categorization at
test\textsuperscript{4}. In contrast, participants in the
\emph{complex-to-simple} curriculum performed more poorly on exceptions
(Log-Odds = \ensuremath{-}1.52 {[}\ensuremath{-}2.41, \ensuremath{-}0.63{]}, \emph{z} = \ensuremath{-}3.35, \emph{p}
\textless{} .001) and did not significantly improve performance on rule
stimuli relative to \emph{no curriculum} (Log-Odds = 0.32 {[}\ensuremath{-}0.28,
0.93{]}, \emph{z} = 1.05, \emph{p} = .296). This differential effect
across rules and exceptions was reflected in a significant curriculum \ensuremath{\times}
stimulus-type interaction (Log-Odds = \ensuremath{-}0.92 {[}\ensuremath{-}1.43, \ensuremath{-}0.42{]}, \emph{z}
= \ensuremath{-}3.57, \emph{p} \textless{} .001). Thus, the \emph{complex-to-simple}
curriculum selectively harmed exception learning, which is consistent
with participants overwriting or catastrophically forgetting exceptions
learned early in training.

Neural network performance mirrors that of humans

Next, we trained a two-layer feedforward neural network (from small
initial weights, \(\sigma\)\, =\,0.01) with 32 ReLU hidden units, one
sigmoid output to perform the R+E task (\textbf{Fig. 1c}). Networks
learned over the same number of trials as humans, and we used
single-trial gradient descent, so that (like humans) networks received
feedback after every trial. Training stimuli were encoded as Boolean
inputs in \{-1,1\}\textsuperscript{3} (\textbf{Fig. 1d}). Specifically,
networks were trained in a yoked manner, with each network exposed to
the exact 240-trial sequence experienced by a single participant in a
given curriculum condition. A key modelling choice was to inject
Gaussian noise into the model hidden activations. This serves two
purposes: it approximates noise in human neural representations and
decision making, and it places the network in a regime where curriculum
effects are expected to arise\textsuperscript{28}. To fit human learning
dynamics we conducted a hyper-parameter sweep over learning rate (from
0.05 to 0.50) and the hidden-activation noise (from 0 to 1.5), and
fitted a global lapse rate. We then selected the configuration that best
reproduced the average human learning accuracy across all conditions,
blocks, and the two stimulus-types. The best globally fitting network
had a learning rate of 0.30, and hidden-activation noise 1.0 and a
global lapse rate of 0.30 (\emph{RMSE} = 0.077).

First, we classify the strategies used by the networks during learning,
using the same approach as for humans (\textbf{Fig. 2b}). Given hidden
activation noise and variability in stimulation sequence not all
networks perform alike. Indeed, the fraction of networks that were
classified under each strategy in each block was markedly similar to
humans (\textbf{Fig. 2a}, lower panels). At test, 51.4\% of networks
trained under the \emph{simple-to-complex} curriculum followed rule plus
exceptions, against 41.3\% under \emph{no curriculum} and 29.2\% under
\emph{complex-to-simple} (\emph{\ensuremath{\chi}\ensuremath{^{2}}}(2, \emph{N} = 207) = 7.39, \emph{p
=} .025, \emph{V} = 0.19), with the proportion following a rule strategy
alone running the other way (19.4\%, 39.7\% and 58.3\%; \emph{\ensuremath{\chi}\ensuremath{^{2}}}(2,
\emph{N} = 207) = 22.87, \emph{p} \textless{} .001, \emph{V} = 0.33). To
assess this human-network state correspondence statistically, we
compared their respective distributions of strategies using a
non-parametric approach. For each curriculum we took the distribution
over the five strategies in each learning block and at test, separately
for humans and networks, we then computed the mean absolute difference
between the distributions of humans and networks for each condition and
for each block. Averaged over the three curricula, this distance was
0.091 when each group of participants was compared with the networks
trained on their own curriculum. We then constructed a null distribution
by randomly assigning each of the 207 networks to one of the three
curriculum conditions, while holding the number of networks per
condition fixed. We then recomputed the matched distance for 5000 such
reassignments. This gave a null centred on 0.120 (95\% range {[}0.112,
0.128{]}). The observed matched distance of 0.091 fell below every one
of the 5000 permutations (\emph{p \textless{}} .001). Thus, the
distribution of network behaviour resembles that of participants trained
under the same curriculum more than those trained under a different one.

Next, we compared the accuracy of networks and humans. Once again, there
was a marked similarity between both human and network learning curves
during training and final performance at test (\textbf{Fig. 3c, d}).
Like human learners, networks trained under the \emph{no curriculum}
condition learned to correctly classify rule stimuli before exceptions
(rule accuracy passed 60\% after a median of 48 trials whereas exception
accuracy took 144 trials and never reached 60\% at all in 5 of 63
networks (paired \emph{t(}57) = 12.01, \emph{p \textless{}} .001;
\textbf{Fig. 3c,} middle). As for humans, exception accuracy first fell
below chance, remaining there from the first trial until trial 139 and
reaching a minimum of 0.440 around trial 70. Networks consistently
showed this effect such that the lowest exception accuracy over the
first half of training averaged 0.342 (SD 0.076) and fell below chance
in 63 of 63 networks (\emph{t(}62) = \ensuremath{-}16.41, \emph{p \textless{}} .001).
Below-chance performance is the signature of all eight stimuli
preferentially being classified by the rule. This pattern of data
implies that networks, like humans, show simplicity biases whereby
gradient-based learning preferentially acquires dominant,
low-dimensional structure before finer-grained distinctions.

Strong similarities between human and network performance were also
observed under the other two curricula. Networks trained with
\emph{simple-to-complex} curricula display a dramatic early drop of
exception accuracy to an average of 0.217, reflecting an even stronger
acquisition of a simple, 1-D rule that similarly misclassifies
exceptions along the rule axis (\textbf{Fig. 3c, right}). However, the
gradually increasing proportion of exceptions drives fast recovery such
that we find improved final exception accuracy relative to \emph{no
curriculum} by the end of training (\textbf{Fig. 3c, center}). However,
in both humans and neural networks this recovery in exception accuracy
is accompanied by some reduction in rule accuracy.

At test, we again see convergence between human and network performance
(\textbf{Fig. 3d}). Networks trained under the \emph{simple-to-complex}
curriculum performed better than those trained without one (Estimate =
0.024 {[}0.007, 0.040{]}, z = 2.82, \emph{p =} .005; network estimates
are differences in accuracy rather than log-odds). Unlike in humans,
however, this benefit was confined to exception stimuli (Estimate =
0.071 {[}0.037, 0.104{]}, z = 4.16, \emph{p \textless{}} .001) and
reversed on rule-consistent stimuli, where networks paid a small but
reliable cost (Estimate = \ensuremath{-}0.024 {[}\ensuremath{-}0.035, \ensuremath{-}0.012{]}, z = \ensuremath{-}4.14,
\emph{p \textless{}} .001). As in humans, networks trained under the
\emph{complex-to-simple} curriculum performed markedly worse on
exceptions (Estimate = \ensuremath{-}0.129 {[}\ensuremath{-}0.162, \ensuremath{-}0.095{]}, z = \ensuremath{-}7.57, \emph{p
\textless{}} .001), while improving slightly on rule stimuli (Estimate =
0.020 {[}0.008, 0.031{]}, z = 3.45, \emph{p \textless{}} .001). This
differential effect across rules and exceptions was reflected in
significant curriculum \ensuremath{\times} stimulus-type interactions in both directions
(\emph{simple-to-complex}: Estimate = 0.094 {[}0.057, 0.131{]}, z =
4.96, \emph{p \textless{}} .001; \emph{complex-to-simple}: Estimate =
\ensuremath{-}0.148 {[}\ensuremath{-}0.185, \ensuremath{-}0.111{]}, z = \ensuremath{-}7.80, \emph{p \textless{}} .001). This
is perhaps due to greater levels of forgetting (catastrophic
interference) in the neural networks.

Because exception frequency differs across conditions in the final
block of training (50\% under \emph{simple-to-complex}, 25\% under
\emph{no curriculum}, 0\% under \emph{complex-to-simple}), the ordering
at test could in principle reflect a recency bias rather than durable
learning. To control for this, we retrained every network with an
identical balanced block appended after block 5 (48 trials at the
\emph{no-curriculum} composition), so that all three conditions end on
the same local statistics. The \emph{simple-to-complex} exception
benefit survives this washout (Estimate = +0.040 {[}0.032, 0.048{]}
against +0.063 {[}0.053, 0.073{]} without it, both \emph{p} $<$ .001),
whereas the \emph{complex-to-simple} exception cost (+0.001
{[}$-$0.007, +0.009{]}, \emph{p} = .73) and the
\emph{simple-to-complex} rule cost ($-$0.001 {[}$-$0.004, +0.001{]},
\emph{p} = .37) do not survive at all (\textbf{Supplementary Fig. 1}).
Recency therefore accounts for the costs that curricula impose,
including the rule cost on which networks and humans diverge, but not
for the observed exception benefit.

Together, these results imply that ordering learning examples from
simple to complex helps both human and artificial learners mitigate
their simplicity biases. In both cases, \emph{simple-to-complex}
curricula successfully layer exception training on earlier rule
acquisition. This effect was not specific to the parameterisation that
best fit human data. When we swept over learning rates and noise levels
and computed the exception benefit for each parameterisation, we found
that it was highly consistent (\textbf{Fig. 3e-f}). However, the
converse is not true; \emph{complex-to-simple} curricula (which train
rules after exceptions) do not preserve initial learning in the same
way. Next, we examine our neural network models to gain a mechanistic
understanding of the benefit of \emph{simple-to-complex} curricula.

\subsubsection{Simple-to-complex curricula facilitate
uncompression}\label{simple-to-complex-curricula-facilitate-uncompression}

Our main prediction is that curricula work by promoting learning along
dimensions that were previously compressed away. If so,
\emph{simple-to-complex} curricula in the R+E task should drive networks
to form a higher-dimensional representation of the stimuli at the end of
training. We test this hypothesis by quantifying the participation ratio
(PR) of the eight stimulus representations in the hidden layer (a
measure of representational dimensionality; \textbf{Fig. 4a}). Final
dimensionality of the stimulus representations ordered as predicted by
the end of training and was higher for \emph{simple-to-complex} compared
to \emph{no curriculum} (2.37 against 2.27 at the end of training;
+0.104 {[}0.046, 0.162{]}, Welch's \emph{t(}127) = 3.52, \emph{p
\textless{}} .001).

To study how network geometry evolves over the course of training under
each curriculum, we projected the hidden representations onto two
task-defined axes (see Methods): a rule axis, the `good' minus `bad'
stimulus centroid for rule stimuli, and an exception axis, the direction
separating the representation of the two exception stimuli. We find that
the curriculum acts on these two axes very differently (\textbf{Fig. 4c,
d}). By the end of training, separation along the exception axis is
substantially greater under the \emph{simple-to-complex} curriculum than
under \emph{no curriculum} (+1.58 {[}0.93, 2.23{]}, Welch's
\emph{t}(122) = 4.82, \emph{p \textless{}} .001), whereas separation
along the rule axis is statistically indistinguishable between them
(+0.19 {[}\ensuremath{-}0.15, 0.52{]}, Welch's \emph{t(}127) = 1.09, \emph{p =} .28).
Any representational expansion is thus specific to the task-relevant
exception representations that had been compressed away early in
training. Visualising these projections confirms this (\textbf{Fig.
4e}). Early in training, after the first block (48 trials) all eight
stimuli are organised almost entirely along the rule axis, with the
exceptions represented as if they were consistent with the rule (which
explains the initial below-chance exception accuracy reported above). By
the end of training the exception representations become separated, and
this effect is stronger for the \emph{simple-to-complex} curriculum than
under \emph{no curriculum}. Across networks, the final participation
ratio predicts final exception accuracy (\emph{r} = 0.70 {[}0.62,
0.76{]}, \emph{p \textless{}} .001, n = 207; \textbf{Fig. 4b}),
confirming that this uncompression connects to model behaviour so that
networks that end up less compressed are those that learn the
exceptions.

\begin{figure}[htbp]
\centering
\includegraphics[width=\linewidth]{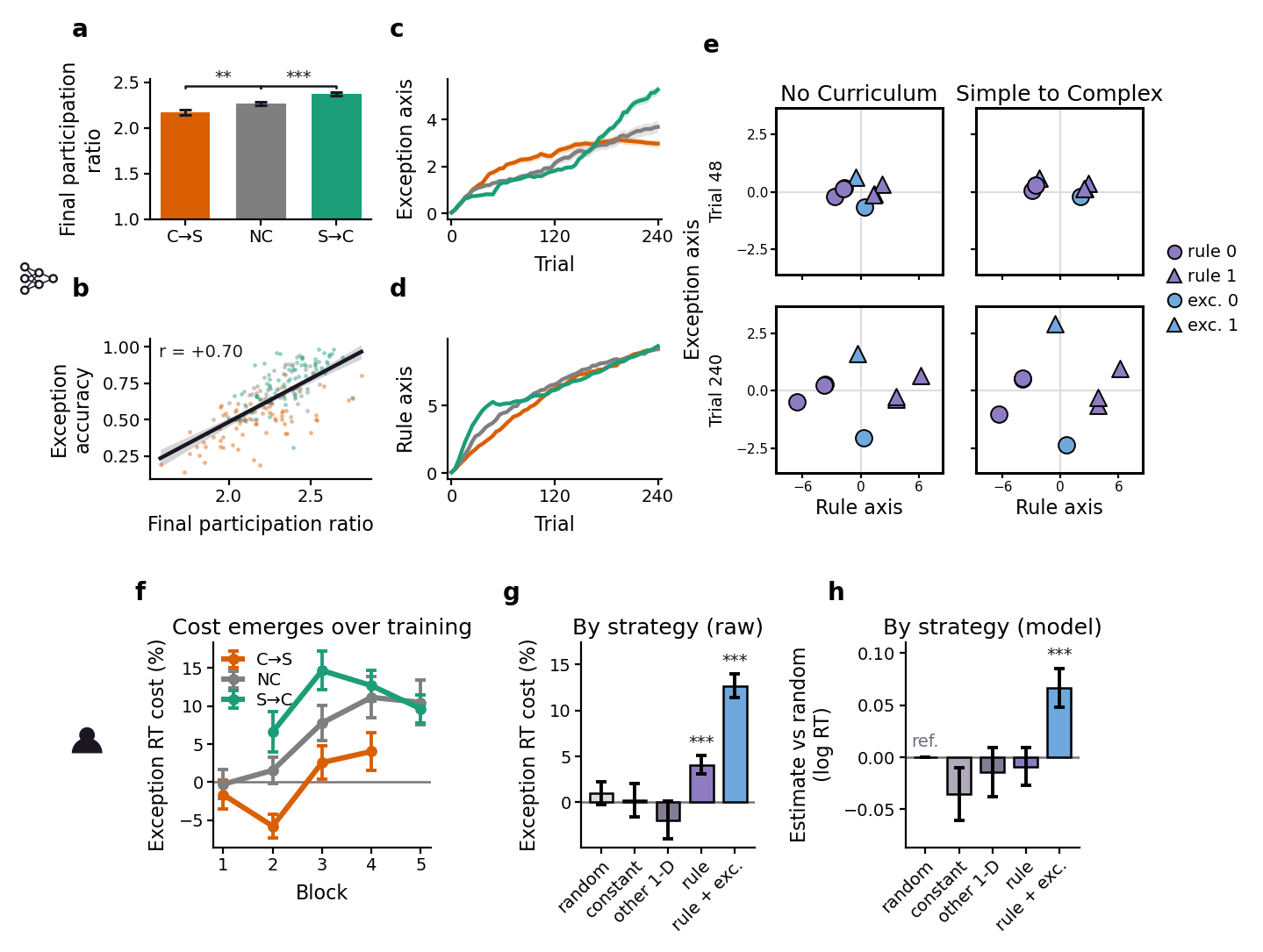}
\caption{\textbf{Uncompression in networks, and the reaction-time
signature of the strategies.} \textbf{a,} Participation ratio of the
eight hidden-layer stimulus representations at the end of training, by
curriculum. \textbf{b,} Final participation ratio against final
exception accuracy, one point per network (\emph{r} = +0.70, \emph{p}
\textless{} .001). \textbf{c, d} Separation of the hidden
representations along the exception axis and the rule axis over
training. The rule axis is the class-1 minus class-0 mean over the six
rule stimuli; the exception axis is the direction separating the two
exceptions, orthogonalised against the rule axis. Both axes have a fixed
sign in every network, so projections can be averaged. \textbf{e,} The
eight stimuli projected onto those axes early in training (trial 48) and
at the end (trial 240), for \emph{no curriculum} and
\emph{simple-to-complex}. Colour denotes stimulus type and shape the
category label, as in Figure 1D. By the end of training the exceptions
are pulled further apart under the curriculum (\emph{simple-to-complex}
minus \emph{no} \emph{curriculum}: +1.58, \emph{t(}122) = 4.82, \emph{p
\textless{}} .001), and the final participation ratio is higher (+0.104,
\emph{t(}127) = 3.52, \emph{p \textless{}} .001). \textbf{f--h} In human
participants, the reaction-time cost of exception stimuli relative to
rule stimuli: over blocks and by curriculum (\textbf{f}), by the
strategy the participant was in (\textbf{g}, raw), and from a mixed
model with participant random effects (\textbf{h}, estimates relative to
random responding). The cost is largest for participants following rule
+ exceptions, as expected if they are retrieving an exception rather
than applying the rule. Stars give \emph{p \textless{}} .05, .01, .001;
error bars \ensuremath{\pm}1 SEM; n = 891 participant-blocks from 207 participants.}
\end{figure}

For humans, we cannot directly measure the dimensionality of task
representations. However, we can analyse reaction times (RTs) to derive
a sensible proxy. If a participant is neglecting the two
exception-relevant features (i.e. those that do not define the rule)
then we should see similar RTs for exception and rule stimuli;
otherwise, exceptions should be slower. We thus computed, for each
participant and block, the difference in mean log RT between exception
and rule trials which we label the \emph{exception RT cost}.

This exception RT cost varies differently over the course of training in
each condition (\textbf{Fig. 4f}). In the no-curriculum condition, the
cost is indistinguishable from zero in block 1 (\ensuremath{-}0.3\%, \emph{t(}62) =
\ensuremath{-}0.15, \emph{p =} .88) and block 2 (+1.5\%, \emph{t(}62) = 0.87, \emph{p
=} .39), but departs from zero from block 3 onwards (+7.8\%,
\emph{t(}62) = 3.46, \emph{p \textless{}} .001) and remains so through
blocks 4 (+11.1\%, \emph{t(}62) = 4.34, \emph{p \textless{}} .001) and 5
(+10.5\%, \emph{t(}62) = 3.75, \emph{p \textless{}} .001). Thus, early
on, participants treat exceptions exactly like rule items, and only
later do they pay a processing cost for them in terms of RT. The
exception RT cost in the curriculum conditions is at least partly driven
by the ratio of rule to exception stimuli, but is comparable in block 3,
where all three conditions present exceptions at an equal rate of 25\%.
Here, we see that the cost is larger under the \emph{simple-to-complex}
curriculum than under \emph{no curriculum} (+14.7\% against +7.8\%; +6.9
points, Welch\textquotesingle s \emph{t}(133) = 2.01, \emph{p =} .046)
and not reliably smaller under \emph{complex-to-simple} (+2.6\%; \ensuremath{-}5.2
points, Welch\textquotesingle s \emph{t}(132) = \ensuremath{-}1.65, \emph{p =} .10).
This ordering matches the strategy distributions in \textbf{Fig. 2a},
and implies that, on average, by block 3 of training, participants are
already ``uncompressing'' the 1D solution that dominates during the
early part of training, and that the \emph{simple-to-complex} curriculum
helps them do so. We also see that the RT cost is greatest when
participants are estimated to be following a rule+exception strategy,
exceeding every other state including the rule-only state (+0.075
{[}0.046, 0.105{]}, z = 5.00, \emph{p \textless{}} .001) validating our
initial strategy inference approach (\textbf{Fig. 4g, h}).

Overall, these results in networks and humans imply that curricula drive
stronger uncompression of exception-relevant dimensions, supporting our
hypothesis and suggesting a further prediction. Curricula should help
most when there is more to unlearn, that is, when the learner begins
from a lower-dimensional, more structured initial representation. In the
next section, we examine if this potential link between initial
representation and curriculum benefit can be manipulated directly in
human learners.

\subsection{\texorpdfstring{Curriculum benefits more for learners with
low-dimensional priors
}{Curriculum benefits more for learners with low-dimensional priors }}\label{curriculum-benefits-more-for-learners-with-low-dimensional-priors}

\begin{figure}[htbp]
\centering
\includegraphics[width=0.85\linewidth]{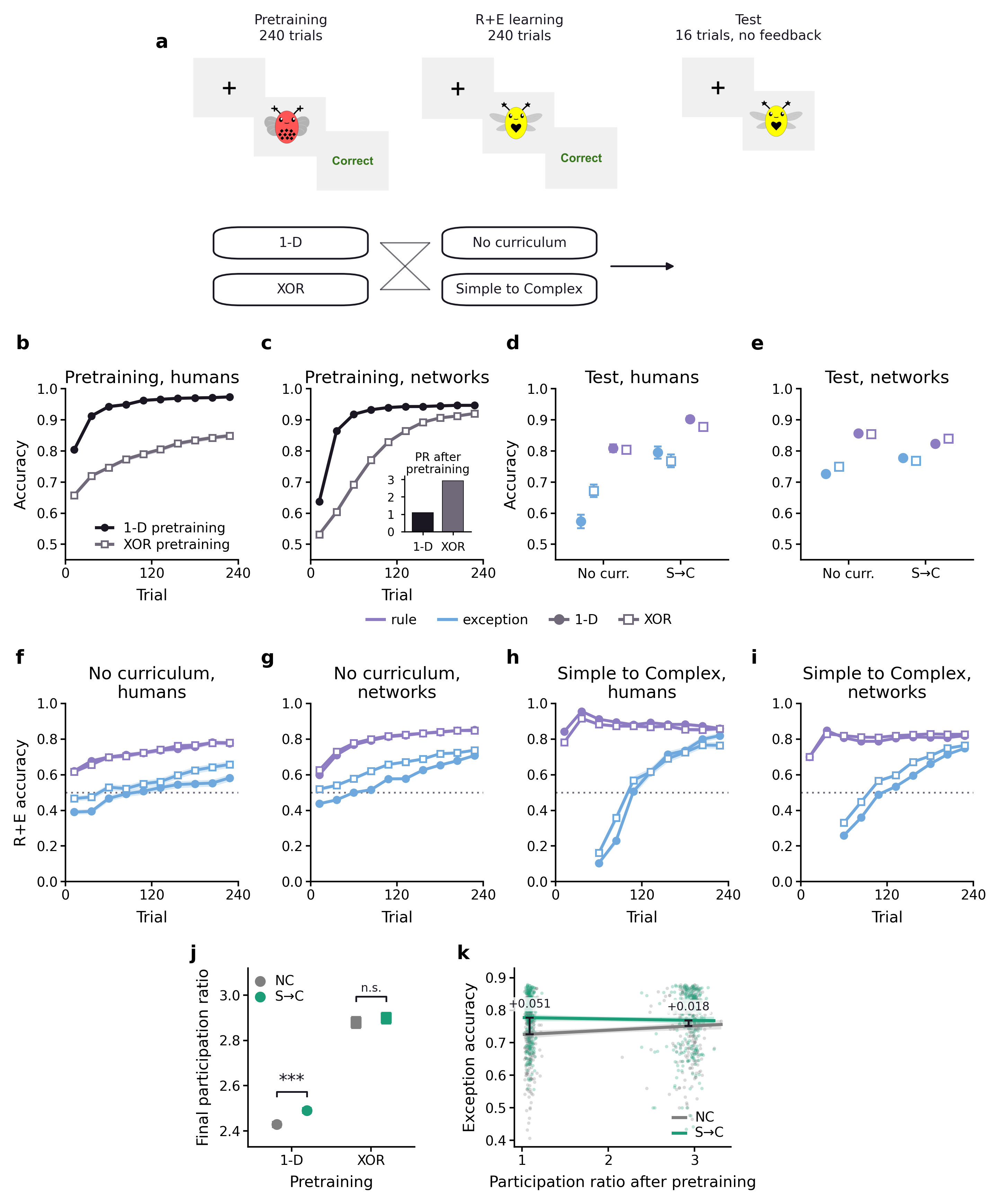}
\caption{\textbf{Pretraining on a 1-D task strengthens simplicity
biases and leads to greater curriculum benefit.} Error bars and shaded
ribbons represent the standard error of the mean (SEM). \textbf{a,} Exp.
2 task design. Participants learn to classify a different set of bugs in
pretraining using a 1-D rule based on a single feature or an XOR rule
over two features. In the learning phase, participants learn the rule
plus exception (R+E) task under a \emph{simple-to-complex} curriculum or
\emph{no curriculum}. Finally, participants complete a test phase where
they classify the stimuli from learning without feedback. \textbf{b, c,}
Pretraining accuracy for participants and for networks, binned into
24-trial bins. One network was trained per participant, on that
participant\textquotesingle s own trial sequence and in their own
physical feature coding, through both phases. Every hyperparameter is
inherited unchanged from the Exp. 1 model; one lapse rate per phase is
the only quantity fitted. Inset: participation ratio of the eight hidden
representations at the end of pretraining, 1.09 after one-dimensional
pretraining against 2.93 after XOR --- the compression manipulation
works on the networks as intended. \textbf{d, e,} Test accuracy for rule
(purple) and exception (blue) stimuli, by curriculum, with pretraining
shown by marker shape. \textbf{f--i} Accuracy over the R+E phase by
curriculum, people and networks. Under \emph{simple-to-complex} no
exception trials occur in the first block, so those bins are empty by
design. \textbf{j,} Participation ratio at the end of R+E learning. The
curriculum raises it for one-dimensionally pretrained networks (+0.062,
\emph{p \textless{}} .001) with no detectable effect after XOR
pretraining (+0.020, \emph{p =} .57). \textbf{k,} Exception accuracy at
test against participation ratio after pretraining, one point per
network, with the curriculum benefit marked at each cluster. The benefit
narrows for the most compressed networks.}
\end{figure}

The results so far suggest that \emph{simple-to-complex} curricula help
learners ``uncompress'' overly simple rules for performing a task. This
allows us to make a new prediction: if we pre-train (or
`meta-train'\textsuperscript{59}) learners on a different task for which
relatively more simple or complex rules apply, then a
\emph{simple-to-complex} curriculum applied to the R+E task should
confer more benefit on the former group (who learn simple rules and thus
compress more). In Exp.2 (\textbf{Fig. 5a)}, we tested this prediction,
in both networks and a new and larger cohort of participants (see
Methods), focussing exclusively on the \emph{no curriculum} and
\emph{simple-to-complex} conditions. In the data that follow, networks
were retrained using hyperparameters obtained from Exp. 1 (width (32),
learning rate (0.3), initialisation scale (\(\sigma\) = 0.01), and
hidden-activation noise (1.0)) directly. The only quantities fit to Exp.
2 are two lapse rates, one per experimental phase; the weights of the
network were carried over from pre-training to the subsequent R+E task.

In Exp. 2, learners first performed a pre-training phase, which was
designed to manipulate the degree of \emph{a priori} representational
compression. One group learned to classify bugs using a simple 1D rule
(e.g. round wings good, double wings bad), whereas the other learned to
apply a higher-dimensional XOR rule defined over two features (e.g.
round wings + heart body or double wings + stripy body good; double
wings + heart body or round wings + stripy body bad; this task requires
learners to attend to at least two dimensions). For Exp.2 we first
report results from neural networks, followed by their corresponding
human participants.

Networks learned both pretraining tasks (\textbf{Fig. 5c}), with
accuracy rising from 0.64 (block 1) to 0.95 (block 10) under the
one-dimensional rule and from 0.53 (block 1) to 0.92 (block 10) under
XOR. Statistically, accuracy on the XOR task thus ended lower (-0.026,
Welch's \emph{t}(538) = -15.48, \emph{p \textless{}} .001). Critically,
in networks we saw that the manipulation had its intended effect on
dimensionality of representations. At the end of pretraining the
participation ratio of the eight hidden representations was 1.09 after
one-dimensional pretraining against 2.93 after XOR (Welch's
\emph{t}(564) = -308, \emph{p \textless{}} .001; \textbf{Fig. 5c}
inset). One-dimensional pretraining leaves the network with a latent
representation that has collapsed to essentially a single dimension,
whereas XOR pretraining induces a more uncompressed, higher dimensional
representation.

Next, we examined subsequent R+E learning. Under \emph{no curriculum},
networks pretrained on the one-dimensional rule reached lower exception
accuracy on the final 24 trials of training than those pretrained on XOR
(0.708 against 0.738; Estimate = \ensuremath{-}0.030 {[}\ensuremath{-}0.053, \ensuremath{-}0.007{]}, Welch's
\emph{t}(516) = \ensuremath{-}2.55, \emph{p =} .011; \textbf{Fig. 5g, i}). In other
words, networks preferred simple solutions, and this was stronger after
1-D pretraining. Critically, however, at test we observed a three-way
interaction between pretraining, curriculum, and stimulus type (Estimate
= 0.052 {[}0.028, 0.077{]}, z = 4.14, \emph{p \textless{}} .001;
\textbf{Fig. 5e}). The curriculum benefit (\emph{simple-to-complex}
relative to \emph{no curriculum}) on exception stimuli was nearly three
times larger after one-dimensional pretraining than after XOR
pretraining (0.051 {[}0.035, 0.066{]}, z = 6.33, \emph{p \textless{}}
.001 against 0.018 {[}0.002, 0.034{]}, z = 2.17, \emph{p =} .030;
difference 0.033 {[}0.010, 0.055{]}, z = 2.87, \emph{p =} .004), with no corresponding advantage on rule stimuli. In neural networks, thus, the
\emph{simple-to-complex} curriculum helps learning most on exception
stimuli when the learner arrives with a low-dimensional prior.
To establish that initial representational change causally mediates
this pretraining effect, we repeated Experiment 2 in neural networks
with the hidden layer frozen at initialisation, so that only the
readout is trained. Because hidden representations do not move, this
yielded networks whose participation ratio was equal in both groups
(3.36 in both) but which still carry their readout weights across the
phase boundary. These networks also displayed a large curriculum
benefit on exception learning, but pretraining condition did not
modulate this effect, with the same benefit after one-dimensional and
XOR pretraining (+0.111 {[}0.083, 0.139{]} against +0.113 {[}0.085,
0.142{]}; difference $-$0.003 {[}$-$0.042, +0.037{]}, \emph{p} = .90)
and no three-way interaction ($-$0.004 {[}$-$0.031, +0.022{]},
\emph{p} = .75; \textbf{Supplementary Fig. 2}). Carrying prior learning
in the readout alone is therefore not sufficient; the effect requires a
representation that prior learning has compressed.

Consistent with the mechanism proposed in Exp. 1, the curriculum also
left simple pretrained networks with a higher participation ratio at the
end of learning (+0.062, Welch's \emph{t}(520) = 3.92, \emph{p
\textless{}} .001), with no detectable effect for networks which
initially received XOR pretraining (+0.019, \emph{p =} .57; \textbf{Fig.
5j, k}). As in Exp. 1, networks that ended learning with a
higher-dimensional representation were those that learned the exceptions
best (\emph{r} = +0.36 within cells, \emph{p \textless{}} .001), and
this relationship was about twice as strong across the four conditions
after one-dimensional than after XOR pretraining (\emph{r =} 0.49 and
0.54 against 0.27 and 0.37).

Humans benefit more from simple-to-complex curricula after 1-D
pretraining

We also collected data from a large cohort of humans performing the same
task. These participants (n = 1027) first completed the same pretraining
phase, designed to manipulate the degree of representational
compression. Pretraining involved a visibly distinct set of bug stimuli
(e.g. red rather than yellow). We reasoned that (like for networks) 1-D
pretraining would encourage participants to compress, whereas XOR
pretraining would require them to retain a higher-dimensional
representation. Participants then learned to perform the same R+E task
as above under either \emph{no curriculum} or a \emph{simple-to-complex}
curriculum. As before, this was followed by a test phase in which
participants classified the learning-phase stimuli without feedback.

During pretraining, accuracy increased over trials in both conditions
(Log-Odds = 1.27 {[}1.21 -- 1.33{]}, \emph{z} = 40.62, \emph{p}
\textless{} .001), but participants are less accurate (Log-Odds = 1.54
{[}1.44 -- 1.65{]}, \emph{z} = 28.72, \emph{p} \textless{} .001) and
improve less (Log-Odds = 0.62 {[}0.56 -- 0.68{]}, \emph{z} = 20.60,
\emph{p} \textless{} .001) when learning an XOR function compared to a
simple function (\textbf{Fig. 5b}). Turning to R+E learning in the
subsequent experimental phase, we then asked whether 1-D pretraining led
to stronger simplicity biases (\textbf{Fig. 5f}). As expected, we see
that participants demonstrated a simplicity bias, performing better
overall on rule than exception stimuli (Log-Odds = -0.47 {[}-0.52 --
-0.43{]}, \emph{z} = -21.39, \emph{p} \textless{} .001). Critically,
participants who received 1-D pretraining were more likely to retain
this simplicity bias than those who received XOR pretraining as
evidenced by a significant pretraining x stimulus type interaction
across all learning blocks (Log-Odds = -0.08 {[}-0.12 -- -0.04{]},
\emph{z} = -3.69, \emph{p} \textless{} .001).

In the learning phase, participants in the \emph{simple-to-complex}
curriculum condition classified exception stimuli less accurately than
rule stimuli (Log-Odds = -1.12 {[}-1.18 -- -1.06{]}, \emph{z} = -36.86,
\emph{p} \textless{} .001) but improved more on exceptions than on rules
across learning (Log-Odds = 0.83 {[}0.78 -- 0.88{]}, \emph{z} = 32.51,
\emph{p} \textless{} .001). Consistent with our predictions, this
differential improvement depended on pretraining. Participants who
received 1-D pretraining showed greater improvement on exceptions
relative to rules than those who received XOR pretraining (Log-Odds =
0.16 {[}0.11 -- 0.21{]}, \emph{z} = 6.59, \emph{p} \textless{} .001). In
other words, the \emph{simple-to-complex} curriculum enabled
participants to recover from the stronger initial simplicity bias
induced by 1-D pretraining.

Next, we examined test-phase performance (\textbf{Fig. 5d}). As in Exp.
1, participants were less accurate on exception stimuli compared to rule
stimuli (Log-Odds = -0.40 {[}-0.50 -- -0.30{]}, z = -7.75, \emph{p
\textless{}} .001). Moreover, participants in the
\emph{simple-to-complex} curriculum once again performed better than
those in \emph{no curriculum} (Log-Odds = 0.56 {[}0.45 -- 0.68{]}, z =
9.71, \emph{p \textless{}} .001). We computed estimated marginal means
contrasts within each stimulus type, which showed the benefit for both
exception stimuli (Log-Odds = 1.35 {[}1.02 -- 1.68{]}, z = 7.93, \emph{p
\textless{}} .001) and rule-consistent stimuli (Log-Odds = 0.92 {[}0.70
-- 1.13{]}, z = 8.15, \emph{p \textless{}} .001). We saw no significant
main effect of pretraining on test performance (Log-Odds = -0.03
{[}-0.14 -- 0.08{]}, z = -0.57, \emph{p =} .57).

Finally, we turned to our key question of interest: do participants who
received 1-D pretraining benefit more from curricula? As in neural
networks, we observed a three-way interaction between pretraining,
curriculum and stimulus type, with comparable functional form (Log-Odds
= 0.10 {[}0.02 -- 0.18{]}, z = 2.33, \emph{p =} .020). The curriculum
benefit was significantly larger for simple- than XOR-pretrained
participants specifically on exception stimuli, mirroring findings in
neural networks. This result supports the hypothesis that the curriculum
facilitates the unlearning of low-dimensional priors in our human
participants.

\subsection{Human curriculum benefits can be predicted by compression
during
pretraining}\label{human-curriculum-benefits-can-be-predicted-by-compression-during-pretraining}

\begin{figure}[htbp]
\centering
\includegraphics[width=\linewidth]{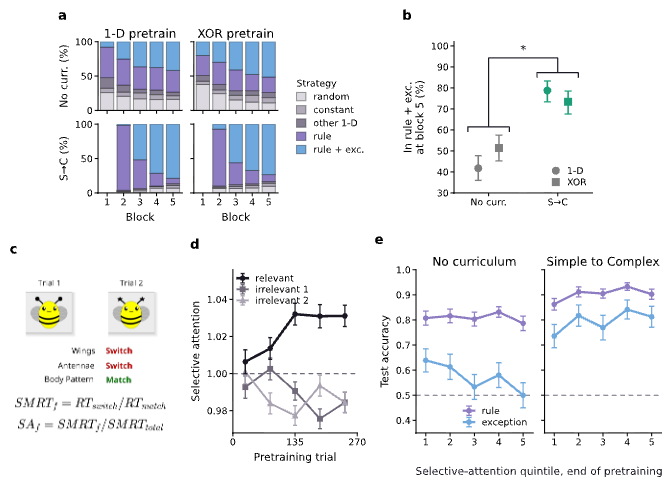}
\caption{\textbf{RT-based measure of compression in pretraining
predicts later curriculum benefit. a.} Proportion of participants in
each strategy over the five R+E learning blocks, split by pretraining
(columns) and curriculum (rows), classified as in Figure 2. \textbf{b.}
Proportion reaching the optimal rule + exceptions strategy by block 5.
With 1-D pretraining fewer participants reach this state without a
curriculum but more with one, giving a significant pretraining \ensuremath{\times}
curriculum interaction (b = \ensuremath{-}0.686, z = \ensuremath{-}2.50, \emph{p =} .012); the
bracket marks that interaction. Error bars are Wilson intervals.
\textbf{c}. We compute a reaction-time measure of selective attention to
each feature dimension during 1-D pretraining. For consecutive trial
pairs, each feature either switches or matches; we define each
feature\textquotesingle s switch-match reaction time (SMRT) as the ratio
of switch to match RTs, where SMRT \textgreater{} 1 indicates attention
to that feature. Selective attention to a feature is its SMRT normalized
by the average SMRT across features. \textbf{d.} During 1-D pretraining,
selective attention to the relevant 1-D feature increases while
selective attention to the two irrelevant features decreases.
Pretraining is split into five bins for computing selective attention,
and error bars are standard errors of participant means. \textbf{e.}
Participants who display greater selective attention to the relevant
feature in 1-D pretraining benefit more from \emph{simple-to-complex}
curricula. Without a curriculum, exception accuracy falls as
participants compress more and under the curriculum it does not.
Selective attention is binned into quintiles for visualization. Error
bars give \ensuremath{\pm}1 SEM.}
\end{figure}

To examine the strategies used after simple and XOR pretraining in more
detail, we adopted the approach taken in Exp.1 to classify learning
strategies (\textbf{Fig. 6a}). Focussing on the final block of training,
we observed that as in Exp.1 more participants adopted the R+E strategy
following a \emph{simple-to-complex} curriculum, but now this effect was
stronger after one dimensional pretraining (41.7\% to 78.8\%, OR = 5.18)
than after XOR pretraining (51.4\% to 73.4\%, OR = 2.61), yielding a
significant pretraining \ensuremath{\times} curriculum interaction in the proportion
classified as rule + exceptions (Likelihood-ratio \ensuremath{\chi}\ensuremath{^{2}} = 6.29, \emph{p =}
.012; \textbf{Fig. 6b}).

Next, we estimated a proxy for representational compression during
pretraining, using a previously-described RT-based measure called the
switch--match reaction time (SMRT)\textsuperscript{60}. This measure
relies on the idea that responses should slow when an attended feature
changes between trials. We divided pretraining into five equally sized
bins and, for each stimulus feature and bin, computed the ratio of
average RTs on trials where the feature changed to those where it did
not (\textbf{Fig. 6c}). SMRT \textgreater{} 1 indicates that
participants slowed on trials where the feature changed (implying they
were attending to it). We normalized each feature's SMRT by the average
SMRT across all three features, yielding a relative measure of selective
attention to that feature (or dimension).

First, we examined how selective attention evolved over the course of
1-D pretraining. As participants learned the task, selective attention
to the relevant 1-D feature was greater overall (Estimate = -0.02
{[}-0.02 -- -0.01{]}, z = -12.04, \emph{p \textless{}} .001) and
increased more over pretraining (Estimate = -0.00 {[}-0.01 -- -0.00{]},
z = -3.73, \emph{p \textless{}} .001) compared to selective attention to
the two irrelevant features (\textbf{Fig. 6d)}. We then asked whether
participants who learned more compressed representations by the end of
1-D pretraining benefitted more from the \emph{simple-to-complex}
curriculum. We found that greater selective attention to the relevant
feature in the last bin of pretraining did not predict overall R+E test
accuracy (Log-Odds = 0.04 {[}-0.12 -- 0.20{]}, \emph{z} = 0.47, \emph{p}
= .64; \textbf{Fig. 6e}). Critically, however, participants who
compressed more by the end of pretraining, benefited more from the
\emph{simple-to-complex} curriculum compared to \emph{no curriculum}
(Log-Odds =0.20 {[}0.04 -- 0.36{]}, \emph{z} = 2.43, \emph{p} = .015).
Together, these findings suggest that the benefit of
\emph{simple-to-complex} curricula for human learners depends on how
much there is to unlearn, i.e. how much relevant features have been
compressed away in prior learning.

\section{Discussion}\label{discussion}

We studied how curriculum learning impacted performance on a classical
categorisation task in which most stimuli followed a simple rule, but
others were exceptions. This task has been studied before, and it is
well known that in the absence of any curriculum, participants tend to
learn rules before exceptions. We replicate this result here. However,
our main question was how and why curricula that sequence rule and
exception stimuli impact learning. Our data suggest that in both humans
and neural networks, simple-to-complex curricula assist learning by
helping people ``uncompress'' -- to escape from an overly simple
solution that only accounts for the rule. We call this phenomenon
``curriculum unlearning''.

We draw upon a literature that has studied the dynamics of learning in
deep networks to understand why rules are learned before
exceptions\textsuperscript{33--38}. It has argued, for example, that
networks prioritise the major modes of the input-output correlation
matrix before minor ones\textsuperscript{33}, or learn linear functions
before more complex polynomials\textsuperscript{37}. Consistent with
this view, we show that networks, like humans, tend to initially learn
1D solutions and only later start to accommodate exceptions. In neural
networks, where we have ready access to the dimensionality and geometry
of neural representation, we see that mastery is accompanied by an
initial compressive phase (in which the rule is acquired) followed by a
gradual uncompression, in which exceptions are acquired. However, the
late phase is not guaranteed; some networks remain trapped in the
minimum formed by initial rule learning. Whilst we did not measure
neural representation change in humans, they showed a strikingly similar
behavioural patterns to neural networks during both training and test,
giving us a strong basis to predict that similar effects would unfold.
Indeed, we already know from neuroimaging studies that the
dimensionality of prefrontal BOLD signals scales with task complexity in
a closely related categorisation task\textsuperscript{54}.

Our main question was how the imposition of a curriculum changes
performance. Curricula that teach regularities before irregularities are
in widespread educational use, such as during second language learning.
We found that people learn faster and more effectively from a curriculum
that prioritises rules before exceptions. Early studies reported that
grouping rule-following stimuli together at the start of each training
block improves learning speed\textsuperscript{4,61} and transfer
success\textsuperscript{62}. A more direct comparison with our work
comes from two reports (using a similar binary categorisation paradigm)
that late introduction of exceptions benefits
performance\textsuperscript{57,58}, and that exception learning may be
driven by hippocampal circuits, which process novel or
out-of-distribution exemplars\textsuperscript{63,64}, and are activated
by exceptions in neuroimaging studies\textsuperscript{65}.
Interestingly, this behavioural effect was observed in adults but not
children. Neural representations in adults have been found to be
intrinsically lower dimensional\textsuperscript{66} (perhaps due to more
extensive training) and so the finding that they benefit more from
curricula is consistent with our theory that curricula help people
``unlearn''.

Our deep learning model offers a candidate explanation for this
curriculum advantage. There is a long tradition of modelling category
learning with connectionist approaches, but standard neural networks
trained with backpropagation were widely believed to be too flexible to
capture human performance on the binary choice employed
here\textsuperscript{50,52}. Nevertheless, our neural network captures
the patterns of learning (during the training phase) and inference
(during the test phase) displayed by humans with quite striking
fidelity. In particular, neural networks show very similar benefits to
humans from simple-to-complex curricula, and similar costs for the
anti-curriculum. This is surprising, because the benefits of curriculum
learning have often been found to be modest or absent in supervised
neural networks\textsuperscript{20}. The clearest theoretically
validated benefits are those for a different training approach (called
``fading'') where exemplars are initially sampled far from the category
boundary, before converging towards it. This approach, in which ``easy''
examples precede ``hard'' ones, benefits humans\textsuperscript{8,9} and
can facilitate learning in neural networks, especially when they are
underparameterised\textsuperscript{28}, due to the geometry of gradient
descent\textsuperscript{26}.

In our R+E task, for both humans and neural networks the curriculum
benefit is realised mainly as a greater probability that learners
transition from simple 1D rules that are underfit to the task (to the
ground truth rule plus exception strategy) during simple-to-complex
curricula. For exceptions, this benefit is very similar between humans
and networks; for rule stimuli, neural networks show a slight
degradation of performance after simple-to-complex training which is
absent in people. This tempers the advantage for the curriculum -- in
fact, in a regime where exceptions are sufficiently rare, it can offset
it entirely. We chose to model learning on this task using an
undifferentiated neural network, allowing us the simplest possible
window into how people perform the task. However, our network lacks
salient architectural motifs, such as the division between hippocampus
and neocortex, which are important for learning in biological
systems\textsuperscript{67}. Coupled with evidence showing that the
hippocampus is key for learning exceptions\textsuperscript{58,65}, it
seems likely that a more comprehensive curriculum benefit will emerge
from models that incorporate this neuroanatomical
detail\textsuperscript{58}.

Studying the dynamics of learning in the neural networks led us to a
computational account of why simple-to-complex curricula benefit
learning. Early in training, learners are prone to adopt overly
simplistic decision strategies that overlook the exceptions. This is one
instance of a canonical phenomenon in cognitive psychology, whereby
people tend to adopt `satisficing' decision strategies that approximate,
but do not fully capture, a more nuanced optimal
policy\textsuperscript{68--70}. The success of the simple-to-complex
depends on its ability to help people escape from these overly simple
solutions. In neural networks, this occurs as the curriculum encourages
exception stimuli to be represented as increasingly separated along a
neural axis which is orthogonal to that for the rule. This theory
predicts that pretraining regimes that encourage relatively simpler,
low-dimensional solutions should create conditions for greater
curriculum benefit, and we show that this is the case for both neural
networks and humans.

We note a number of limitations to our study. Firstly, our neural
network models, like many other computational tools, offer an overly
simplistic account of human learning. Although it is surprising that
network models capture the overall learning dynamics very faithfully,
there are still some important ways in which they misfit -- they show
hallmarks of forgetting for rule stimuli, which humans do not, and
although they mimic patterns of human curriculum benefit, those benefits
in the networks are overall more modest. However, the simplicity of
neural networks models is also their virtue, as it offers us an elegant
theory of the phenomenon we set out to study, and an unclouded window
into how representational geometry and dimensionality change over the
course of learning.

\section*{References}\label{references}

1. Thornton, A. \& McAuliffe, K. Teaching in Wild Meerkats.
\emph{Science} \textbf{313}, 227--229 (2006).

2. Musgrave, S. \emph{et al.} Teaching varies with task complexity in
wild chimpanzees. \emph{Proc. Natl. Acad. Sci.} \textbf{117}, 969--976
(2020).

3. Bruner, J. S. \emph{The Process of Education}. (Harvard university
press, Cambridge (Mass.), 1977).

4. Mathy, F. \& Feldman, J. A rule-based presentation order facilitates
category learning. \emph{Psychon. Bull. Rev.} \textbf{16}, 1050--1057
(2009).

5. Mi, Q. \& Summerfield, C. Human curriculum learning of a cue
combination task. \emph{Nat. Hum. Behav.}
https://doi.org/10.1038/s41562-026-02452-1 (2026)
doi:10.1038/s41562-026-02452-1.

6. Antzoulatos, E. G. \& Miller, E. K. Increases in functional
connectivity between prefrontal cortex and striatum during category
learning. \emph{Neuron} \textbf{83}, 216--25 (2014).

7. Elio, R. \& Anderson, J. R. The effects of information order and
learning mode on schema abstraction. \emph{Mem Cogn.} \textbf{12},
20--30 (1984).

8. Pashler, H. \& Mozer, M. C. When does fading enhance perceptual
category learning? \emph{J. Exp. Psychol. Learn. Mem. Cogn.}
\textbf{39}, 1162--1173 (2013).

9. Hornsby, A. N. \& Love, B. C. Improved classification of mammograms
following idealized training. \emph{J. Appl. Res. Mem. Cogn.}
\textbf{3}, 72--76 (2014).

10. Strittmatter, Y., Mannelli, S. S., Ruiz-Garcia, M., Musslick, S. \&
Spitzer, M. Curriculum learning in humans and neural networks. Preprint
at https://doi.org/10.31234/osf.io/a9kjb\_v2 (2025).

11. Volk, C., Pack, C. C. \& Bakhtiari, S. The curriculum effect in
visual learning: the role of readout dimensionality. Preprint at
https://doi.org/10.1101/2025.09.19.677409 (2025).

12. Ahissar, M. \& Hochstein, S. Task difficulty and the specificity of
perceptual learning. \emph{Nature} \textbf{387}, 401--406 (1997).

13. Lawrence, D. H. The transfer of a discrimination along a continuum.
\emph{J. Comp. Physiol. Psychol.} \textbf{45}, 511--516 (1952).

14. Wulf, G. \& Shea, C. H. Principles derived from the study of simple
skills do not generalize to complex skill learning. \emph{Psychon. Bull.
Rev.} \textbf{9}, 185--211 (2002).

15. Skinner, B. F. How to Teach Animals. in \emph{Cumulative record
(enlarged ed.).} 412--419 (Appleton-Century-Crofts, East Norwalk, 1961).
doi:10.1037/11324-031.

16. Boundey, A., Nikitins, F. \& Fernandez, E. J. An examination of
husbandry training in zoos: A scoping review. \emph{Appl. Anim. Behav.
Sci.} \textbf{292}, 106805 (2025).

17. Richards, B. A. \emph{et al.} A deep learning framework for
neuroscience. \emph{Nat Neurosci} \textbf{22}, 1761--1770 (2019).

18. Saxe, A., Nelli, S. \& Summerfield, C. If deep learning is the
answer, what is the question? \emph{Nat. Rev. Neurosci.} \textbf{22},
55--67 (2021).

19. Marblestone, A. H., Wayne, G. \& Kording, K. P. Toward an
Integration of Deep Learning and Neuroscience. \emph{Front Comput
Neurosci} \textbf{10}, 94 (2016).

20. Wu, X., Dyer, E. \& Neyshabur, B. When Do Curricula Work? Preprint
at http://arxiv.org/abs/2012.03107 (2021).

21. Charpentier, L. \emph{et al.} Findings of the Third BabyLM
Challenge: Accelerating Language Modeling Research with Cognitively
Plausible Data. in \emph{Proceedings of the First BabyLM Workshop}
399--420 (Association for Computational Linguistics, Suzhou, China,
2025). doi:10.18653/v1/2025.babylm-main.28.

22. Lee, J. H., Mannelli, S. S. \& Saxe, A. Why Do Animals Need Shaping?
A Theory of Task Composition and Curriculum Learning. Preprint at
https://doi.org/10.48550/ARXIV.2402.18361 (2024).

23. Krueger, K. A. \& Dayan, P. Flexible shaping: how learning in small
steps helps. \emph{Cognition} \textbf{110}, 380--94 (2009).

24. Elman, J. L. Learning and development in neural networks: the
importance of starting small. \emph{Cognition} \textbf{48}, 71--99
(1993).

25. Khajehabdollahi, S. \emph{et al.} Emergent mechanisms for long
timescales depend on training curriculum and affect performance in
memory tasks. https://doi.org/10.48550/ARXIV.2309.12927 (2023)
doi:10.48550/ARXIV.2309.12927.

26. Weinshall, D. \& Amir, D. Theory of Curriculum Learning, with Convex
Loss Functions. \emph{ArXiv181203472 Cs Stat}
http://arxiv.org/abs/1812.03472 (2018).

27. Hacohen, G. \& Weinshall, D. On The Power of Curriculum Learning in
Training Deep Networks. Preprint at
https://doi.org/10.48550/arXiv.1904.03626 (2019).

28. Mannelli, S. S., Ivashynka, Y., Saxe, A. \& Saglietti, L. Tilting
the Odds at the Lottery: the Interplay of Overparameterisation and
Curricula in Neural Networks. Preprint at
http://arxiv.org/abs/2406.01589 (2024).

29. Saglietti, L., Mannelli, S. S. \& Saxe, A. An Analytical Theory of
Curriculum Learning in Teacher-Student Networks.
https://doi.org/10.48550/ARXIV.2106.08068 (2021)
doi:10.48550/ARXIV.2106.08068.

30. Goodfellow, I., Bengio, Y. \& Courville, A. \emph{Deep Learning}.
(MIT Press, 2016).

31. LeCun, Y., Bengio, Y. \& Hinton, G. Deep learning. \emph{Nature}
\textbf{521}, 436--44 (2015).

32. Flesch, T., Juechems, K., Dumbalska, T., Saxe, A. \& Summerfield, C.
Orthogonal representations for robust context-dependent task performance
in brains and neural networks. \emph{Neuron} S0896627322000058 (2022)
doi:10.1016/j.neuron.2022.01.005.

33. Saxe, A. M., McClelland, J. L. \& Ganguli, S. A mathematical theory
of semantic development in deep neural networks. \emph{Proc Natl Acad
Sci U A} \textbf{116}, 11537--11546 (2019).

34. Rubruck, J., Bauer, J. P., Saxe, A. \& Summerfield, C. Early
learning of the optimal constant solution in neural networks and humans.
Preprint at https://doi.org/10.48550/ARXIV.2406.17467 (2024).

35. Tiwari, R. \& Shenoy, P. Overcoming Simplicity Bias in Deep Networks
using a Feature Sieve. Preprint at
https://doi.org/10.48550/ARXIV.2301.13293 (2023).

36. Nakkiran, P. \emph{et al.} SGD on Neural Networks Learns Functions
of Increasing Complexity. Preprint at
https://doi.org/10.48550/ARXIV.1905.11604 (2019).

37. Rahaman, N. \emph{et al.} On the Spectral Bias of Neural Networks.
https://doi.org/10.48550/ARXIV.1806.08734 (2018)
doi:10.48550/ARXIV.1806.08734.

38. Bhattamishra, S., Patel, A., Kanade, V. \& Blunsom, P. Simplicity
Bias in Transformers and their Ability to Learn Sparse Boolean
Functions. Preprint at https://doi.org/10.48550/ARXIV.2211.12316 (2022).

39. Shwartz-Ziv, R. \& Tishby, N. Opening the Black Box of Deep Neural
Networks via Information. Preprint at
https://doi.org/10.48550/arXiv.1703.00810 (2017).

40. Schuck, N. W. \emph{et al.} Medial Prefrontal Cortex Predicts
Internally Driven Strategy Shifts. \emph{Neuron} \textbf{86}, 331--340
(2015).

41. L\"{o}we, A. T. \emph{et al.} Regularised neural networks mimic human
insight. Preprint at http://arxiv.org/abs/2302.11351 (2023).

42. Sadtler, P. T. \emph{et al.} Neural constraints on learning.
\emph{Nature} \textbf{512}, 423--6 (2014).

43. Oby, E. R. \emph{et al.} New neural activity patterns emerge with
long-term learning. \emph{Proc. Natl. Acad. Sci.} \textbf{116},
15210--15215 (2019).

44. Shepard, R. N., Hovland, C. I. \& Jenkins, H. M. Learning and
memorization of classifications. \emph{Psychol. Monogr. Gen. Appl.}
\textbf{75}, 1--42 (1961).

45. Feldman, J. Minimization of Boolean complexity in human concept
learning. \emph{Nature} \textbf{407}, 630--633 (2000).

46. Johansen, M. Are there representational shifts during category
learning? \emph{Cognit. Psychol.} \textbf{45}, 482--553 (2002).

47. Smith, J. D. \& Minda, J. P. Prototypes in the mist: The early
epochs of category learning. \emph{J. Exp. Psychol. Learn. Mem. Cogn.}
\textbf{24}, 1411--1436 (1998).

48. Nosofsky, R. M., Palmeri, T. J. \& McKinley, S. C.
Rule-plus-exception model of classification learning. \emph{Psychol.
Rev.} \textbf{101}, 53--79 (1994).

49. Anderson, J. R. \& Betz, J. A hybrid model of categorization.
\emph{Psychon. Bull. Rev.} \textbf{8}, 629--647 (2001).

50. Love, B. C., Medin, D. L. \& Gureckis, T. M. SUSTAIN: A Network
Model of Category Learning. \emph{Psychol. Rev.} \textbf{111}, 309--332
(2004).

51. Pothos, E. M. \& Chater, N. Categorization by simplicity: a minimum
description length approach to unsupervised clustering. in
\emph{Similarity and Categorization} (eds Hahn, U. \& Ramscar, M.)
51--72 (Oxford University PressOxford, 2001).
doi:10.1093/acprof:oso/9780198506287.003.0004.

52. Kruschke, J. K. ALCOVE: An exemplar-based connectionist model of
category learning. \emph{Psychol. Rev.} \textbf{99}, 22--44 (1992).

53. Feldman, J. The Simplicity Principle in Human Concept Learning.
\emph{Curr. Dir. Psychol. Sci.} \textbf{12}, 227--232 (2003).

54. Mack, M. L., Preston, A. R. \& Love, B. C. Ventromedial prefrontal
cortex compression during concept learning. \emph{Nat. Commun.}
\textbf{11}, 46 (2020).

55. Carcassi, F. \& Szymanik, J. Neural Networks Track the Logical
Complexity of Boolean Concepts. \emph{Open Mind} \textbf{6}, 132--146
(2022).

56. Mingard, C. \emph{et al.} Neural networks are a priori biased
towards Boolean functions with low entropy. Preprint at
https://doi.org/10.48550/ARXIV.1909.11522 (2019).

57. Song, J. \emph{et al.} Category learning in adults but not children
benefits from delayed introduction of exceptions. \emph{JoCN Forum}
https://doi.org/10.21428/8e6ba8ef.a059aed8 (2026)
doi:10.21428/8e6ba8ef.a059aed8.

58. Heffernan, E. M., Schlichting, M. L. \& Mack, M. L. Learning
exceptions to the rule in human and model via hippocampal encoding.
\emph{Sci. Rep.} \textbf{11}, 21429 (2021).

59. Binz, M. \emph{et al.} Meta-Learned Models of Cognition. Preprint at
http://arxiv.org/abs/2304.06729 (2023).

60. Collins, A. G. E. The Cost of Structure Learning. \emph{J. Cogn.
Neurosci.} \textbf{29}, 1646--1655 (2017).

61. Mezzadri, G., Reynaud-Bouret, P., Lalo\"{e}, T. \& Mathy, F.
Investigating interactions between types of order in categorization.
\emph{Sci. Rep.} \textbf{12}, 21625 (2022).

62. Mathy, F. \& Feldman, J. The Influence of Presentation Order on
Category Transfer. \emph{Exp. Psychol.} \textbf{63}, 59--69 (2016).

63. Sun, W., Advani, M., Spruston, N., Saxe, A. \& Fitzgerald, J. E.
\emph{Organizing Memories for Generalization in Complementary Learning
Systems}. http://biorxiv.org/lookup/doi/10.1101/2021.10.13.463791 (2021)
doi:10.1101/2021.10.13.463791.

64. Strange, B. A., Duggins, A., Penny, W., Dolan, R. J. \& Friston, K.
J. Information theory, novelty and hippocampal responses: unpredicted or
unpredictable? \emph{Neural Netw} \textbf{18}, 225--30 (2005).

65. Davis, T., Love, B. C. \& Preston, A. R. Learning the Exception to
the Rule: Model-Based fMRI Reveals Specialized Representations for
Surprising Category Members. \emph{Cereb. Cortex} \textbf{22}, 260--273
(2012).

66. Busch, E. L. \& Turk-Browne, N. B. Developmental tuning of
functional manifold dimensionality across the human brain. Preprint at
https://doi.org/10.64898/2026.07.24.740635 (2026).

67. McClelland, J. L., McNaughton, B. L. \& O'Reilly, R. C. Why there
are complementary learning systems in the hippocampus and neocortex:
insights from the successes and failures of connectionist models of
learning and memory. \emph{Psychol Rev} \textbf{102}, 419--57 (1995).

68. Lieder, F. \& Griffiths, T. L. Resource-rational analysis:
Understanding human cognition as the optimal use of limited
computational resources. \emph{Behav. Brain Sci.} \textbf{43}, e1
(2020).

69. Shah, A. K. \& Oppenheimer, D. M. Heuristics made easy: An
effort-reduction framework. \emph{Psychol. Bull.} \textbf{134}, 207--222
(2008).

70. Gigerenzer, G. \& Gaissmaier, W. Heuristic decision making.
\emph{Annu Rev Psychol} \textbf{62}, 451--82 (2011).

71. Bates, D., M\"{a}chler, M., Bolker, B. \& Walker, S. Fitting Linear
Mixed-Effects Models Using \textbf{lme4}. \emph{J. Stat. Softw.}
\textbf{67}, (2015).

72. Barr, D. J., Levy, R., Scheepers, C. \& Tily, H. J. Random effects
structure for confirmatory hypothesis testing: Keep it maximal. \emph{J.
Mem. Lang.} \textbf{68}, 255--278 (2013).

73. Lenth, R. V. \& Piaskowski, J. emmeans: Estimated Marginal Means,
aka Least-Squares Means. 2.0.4
https://doi.org/10.32614/CRAN.package.emmeans (2017).

\section{Methods}\label{methods}
\subsection{\texorpdfstring{Experiment 1
}{Experiment 1 }}\label{experiment-1}

\subsubsection{Participants}\label{participants}

207 participants completed Exp. 1. They were between the ages of 18 and
40 and were recruited from the crowd working site Prolific. All
participants provided informed consent before beginning the study, and
procedures were reviewed by the local ethics committee (protocol
R50750/RE010). Only participants who completed the full study and missed
fewer than four catch trials were included in analyses. The study took
approximately 20 minutes to complete, and participants were paid \pounds{}4 plus
a performance bonus per correct response, up to a maximum of \pounds{}1.
Participants were randomly assigned with equal probability to one of
three curriculum conditions: "simple-to-complex," "complex-to-simple,"
or "no curriculum." Of the 207 included participants, 72 learned under
the simple-to-complex curriculum, 72 under complex-to-simple, and 63
under no curriculum.

\subsubsection{Experimental Procedure}\label{experimental-procedure}

Participants completed a category learning task, where they were asked
to classify bug stimuli as either good bugs or bad bugs. Each stimulus
could be identified by the combination of three Boolean features (wing
shape, body pattern, antennae type), resulting in 8 unique stimuli. The
eight stimuli followed a family-resemblance structure: a simple rule
based on a single feature correctly classified six of the eight stimuli,
while the remaining two stimuli were exceptions that violated the rule
(mirroring the R+E structure of the network model\textquotesingle s
task, described below). The mapping between the three physical features
and the three Boolean dimensions, and the assignment of feature values
to 0/1, was randomized independently for each participant, so that no
single physical feature was diagnostic across the sample. The assignment
of the "A" and "L" keys to "good bug" and "bad bug" responses, and the
left/right screen position of the two response options, were also
independently randomized per participant.

On each learning trial, participants saw a bug and had up to 10 seconds
to use the \textquotesingle A\textquotesingle{} or
\textquotesingle L\textquotesingle{} keys to classify it as good or bad.
They then received feedback on their choice for 2 seconds. Participants
completed 5 blocks of learning with 48 bug trials each. Each learning
block also included 6 catch trials designed to detect inattentive
participants, in which a cartoon green bee was shown, and participants
were instructed to press the spacebar. Participants who missed 4 or more
catch trials, or who timed out on more than 5 trials, had their task
automatically ended and were excluded from analysis.

In the simple-to-complex curriculum, the five learning blocks progressed
from containing no exception trials (Block 1: 48 rule-consistent trials,
0 exceptions) to an equal mix of rule and exception trials (Block 5: 24
rule-consistent, 24 exception trials), increasing in even steps across
blocks (Block 2: 6 exceptions {[}12.5\%{]}; Block 3: 12 exceptions
{[}25\%{]}; Block 4: 18 exceptions {[}37.5\%{]}). In the
complex-to-simple curriculum, this same sequence of five blocks was
presented in reverse order. In the no curriculum condition, all five
learning blocks used a fixed composition of 12 exception trials each
(25\%), matching the average exception rate of the two curriculum
conditions.

After the learning phase, participants completed a short test phase,
where they saw each of the eight unique bug stimuli twice, in random
order, and were asked to classify them without feedback.

\subsection{Experiment 2}\label{experiment-2}

A total of 1027 participants completed our study across 5 separate data
collection sessions and were included in analyses. The study took
approximately 40 minutes to complete, and participants were recruited in
the same manner as in Experiment 1. Only participants who completed the
full study and were not excluded for inattentiveness were included in
analyses. Sample size and exclusion criteria were determined a priori
for each data collection session. Participants were randomly assigned
with equal probability to one of two pretraining conditions (simple or
XOR) and one of two curriculum conditions (simple-to-complex or no
curriculum) in a between-subjects design. Of the 1027 included
participants, 266 received 1-D pretraining and no curriculum, 259
received 1-D pretraining and simple-to-complex curriculum, 254 received
XOR pretraining and no curriculum, and 248 received XOR pretraining and
simple-to-complex-curriculum.

\subsubsection{Experimental Procedure}\label{experimental-procedure-1}

\textbf{Pretraining.} Participants learned about a set of bug stimuli
that could be identified by the same three Boolean features as the main
task (wing shape, body pattern, antennae type) but realized in a body
color (red or blue) and set of feature values not otherwise used in the
experiment. Depending on the assigned pretraining condition, the correct
category was determined either by a simple rule based on a single
feature ("simple" pretraining) or by the exclusive-or (XOR) of two
features ("XOR" pretraining), with the third feature irrelevant in both
cases. The mapping between physical features and Boolean dimensions was
randomized independently per participant, as in the main task.
Pretraining consisted of a fixed sequence of 270 trials organized into
30 blocks of 9: within each block, 8 trials presented one of the 8
pretraining stimuli (sampled with replacement) and 1 trial, at a
randomized position within the block, was a catch trial. Feedback was
error-corrective: an incorrect response was immediately followed by
re-presentation of the same stimulus on the next trial, repeating until
the participant responded correctly, after which a new stimulus was
sampled; correct responses and catch trials were always followed by a
new stimulus draw. As in the main task, participants had up to 10
seconds to respond on each trial.

\textbf{Learning phase.} Participants then learned about a new set of
bug stimuli with a different body color and a new set of feature values
not seen during pretraining (the color used for pretraining and the
color used for the learning/test phases were counterbalanced across
participants). The learning-phase stimuli followed the same
6-rule/2-exception structure as Exp. 1. As in Exp. 1, there were 5
learning blocks of 48 learning trials and 6 catch trials each (54
trials/block), with the proportion of exception trials increasing across
blocks (0\%, 12.5\%, 25\%, 37.5\%, 50\%) under the simple-to-complex
curriculum or fixed at 25\% throughout under no curriculum (Exp. 2 did
not include a complex-to-simple condition). Learning blocks were
presented consecutively with no on-screen break or signal marking the
transition between blocks. Participants then completed a test phase
identical in structure to Exp. 1: the eight learning-phase stimuli were
each presented twice (16 trials total), in random order and without
feedback.

\textbf{Data collection sessions.} Data collection sessions 1--3
(combined N = 628 {[}96, 116, and 416, respectively{]}) used one task
implementation. Catch trials in these sessions used a distinctive green
cartoon bee (shared across both pretraining colors) that was visually
dissimilar from every classification stimulus; participants were
automatically excluded from further participation in real time if they
missed 4 or more catch trials or timed out on more than 5 trials.
Participants were paid \pounds{}6 plus a performance bonus per correct response,
up to a maximum of \pounds{}2.

Data collection sessions 4--5 (combined N = 399 {[}78 and 321,
respectively{]}) modified the procedure in three ways relative to
sessions 1--3, intended to increase compliance and better detect
inattentive participants. First, the catch-trial stimulus was changed
from the distinctive green bee to a color-matched "unusual bug" composed
of a novel combination of wing, body-pattern, and antennae features not
used by any other stimulus in the experiment (previewed to participants
on an instruction screen before each phase began), making catch trials
substantially harder to pass by inattentive or non-deliberate
responding. There was no longer a hard, real-time cutoff for missed
catch trials (only excessive timeouts, \textgreater5, ended the task
automatically); catch-trial performance was instead used as a post-hoc
exclusion criterion during analysis. Second, feedback duration was made
asymmetric: 1.5 seconds following correct responses but 4 seconds
following incorrect responses (versus a flat 2 seconds in sessions
1--3), giving participants more time to process error feedback. Third,
participants were paid a flat fee of \pounds{}9 with no performance bonus.

\subsection{Analysis Approach}\label{analysis-approach}

We used the \textquotesingle lme4\textquotesingle{}
package\textsuperscript{71} in R to fit mixed-effects models to our
data. Continuous variables were z-scored prior to model-fitting.
Categorical variables were coded using sum contrasts, with one
exception: the three-level curriculum factor in Exp. 1 was coded using
treatment (dummy) contrasts with "No Curriculum" as the reference level,
so that model coefficients directly reflect the comparison of each
curriculum condition against the no curriculum control. Trials in which
participants failed to respond within the allotted time limit were
excluded from analyses.

We began by fitting models that included random intercepts for each
participant and random slopes for every within-participant predictor and
their interactions\textsuperscript{72}. When models failed to converge,
we pruned correlations between random slopes and intercepts, followed by
interactions between random slopes, followed by random slopes
themselves. We assessed the significance of fixed effects using Wald
tests. Effects of curriculum separately on exception and rule stimuli
were computed as model-based contrasts (emmeans
package\textsuperscript{73}) and tested with Wald \emph{z}-tests.

\textbf{Switch--match reaction time (SMRT).} SMRT was computed from the
pretraining phase only. Each trial was paired with the immediately
preceding trial within the same block, and for each of the three
features the transition was labeled a \emph{match} (same feature value
as the previous trial) or a \emph{switch} (different value). We excluded
the first trial of each block, catch and timed-out trials, any pair that
was not strictly consecutive (e.g., because a catch or timed-out trial
intervened), and any pair whose preceding trial was answered
incorrectly, because error-corrective feedback re-presented the
identical stimulus on the next trial. Pretraining was divided into five
bins of 54 trials. For each participant, bin, and feature, we computed
the mean RT on switch trials and on match trials, and SMRT was their
ratio (switch/match). To obtain a relative measure of selective
attention, each feature's SMRT was divided by the mean SMRT across the
three features within the same participant and bin, so that values above
1 indicate greater slowing when that feature changes than when the other
features change.

\subsection{Strategy classification model
fitting}\label{strategy-classification-model-fitting}

For each participant and block we inferred which of a small set of
decision policies best described their responses. We considered nine
deterministic policies: the correct rule (classify by the relevant
feature); the rule plus exceptions (the correct rule with the two
exception stimuli reversed, i.e. the optimal policy); the relevant
feature with inverted sign; each of the two irrelevant features in
either sign (four policies); and the two constant responses ("always
good", "always bad"). To these we added a tenth candidate, random
responding at p = 0.5.

Each deterministic policy was fitted with a single free parameter, a
lapse rate \ensuremath{\varepsilon} estimated by maximum likelihood as the observed
disagreement rate between the policy\textquotesingle s predictions and
the participant\textquotesingle s responses, clipped to {[}10\ensuremath{^{-6}}, 0.5{]}.
Candidates were compared by BIC, \ensuremath{-}2 log L + log(N) for the deterministic
policies and \ensuremath{-}2 log L with no penalty for the random null, where N is
the number of trials in the block. Leaving the random null unpenalised
makes the classifier conservative and a structured policy is adopted
only if it earns its free parameter, so ambiguous blocks fall to
"random" rather than to a weakly-supported strategy. The nine
deterministic policies were collapsed to four reported labels, giving
five in total: rule; rule + exceptions; other 1-D rule (the inverted
rule and the four irrelevant-feature policies); constant (the two
constant responses); and random. A block was classified only if the
stimuli actually presented in it distinguish rule from rule + exceptions
--- that is, only if at least one exception stimulus occurred. This
excludes block 1 of the simple-to-complex condition and block 5 of the
complex-to-simple condition, which contain no exception trials.

Networks were classified with the identical procedure. Because the
classifier operates on discrete responses, we sampled binary responses
from each network\textquotesingle s noise-marginalised accuracy through
the group lapse rate, at the human trial counts (six presentations of
each stimulus per learning block, two per stimulus at test), and passed
these to the same routine. Classification of the test phase rests on 16
trials against 48 per learning block and is correspondingly less
powered.

\subsection{Neural Networks}\label{neural-networks}

\subsubsection{Stimuli and yoked
training}\label{stimuli-and-yoked-training}

Networks received the same three binary stimulus features available to
participants, coded as \ensuremath{\pm}1, so that the eight stimuli are the corners of
a cube in \ensuremath{\mathbb{R}}\ensuremath{^{3}}. The rule was defined on the second feature, and the two
exception stimuli were those with the first feature present and the
third absent.

Rather than training networks on idealised trial sequences, we trained
one network per participant on that participant\textquotesingle s own
sequence of trials: 207 networks in Exp. 1 and 1027 in Exp. 2. Each
network received exactly the stimuli its matched participant received,
in the same order, with the same curriculum assignment, and with
supervised feedback after every trial. Participants who timed out on a
trial contribute a correspondingly shorter sequence; sequences were
padded to a common length and masked so that padded positions contribute
no gradient.

\subsubsection{Architecture and hyperparameter
search}\label{architecture-and-hyperparameter-search}

Each network was a two-layer feedforward network: three inputs, H = 32
ReLU hidden units, and a single sigmoid output. Weights and biases were
initialised i.i.d. Gaussian with scale 0.01. Networks were trained by
online stochastic gradient descent on the binary cross-entropy loss, one
trial at a time, with learning rate 0.30.

Gaussian noise with \ensuremath{\sigma} = 1.0 was added to the hidden activations after
the ReLU on every trial, so that h = ReLU(Wx + b) + \ensuremath{\varepsilon}, \ensuremath{\varepsilon}
\textasciitilde{} \ensuremath{\mathcal{N}}(0, \ensuremath{\sigma}\ensuremath{^{2}}I), drawn independently each trial. Responses
were generated through a lapse observation model, p\ensuremath{\prime} = (1 \ensuremath{-} \ensuremath{\varepsilon})p + \ensuremath{\varepsilon}/2,
with a single group-level lapse \ensuremath{\varepsilon} = 0.30. The lapse is applied to
responses only and never enters the weight update. We fixed the
initialisation scale at 0.01 (the rich, feature-learning regime). We
then swept learning rate \ensuremath{\in} \{0.05, 0.10, 0.20, 0.30, 0.50\} \ensuremath{\times} hidden
noise \ensuremath{\in} \{0, 0.25, 0.50, 0.75, 1.00, 1.25, 1.50\}, 35 configurations at
32 hidden units with three seeds each. Configurations were scored by
RMSE against the human group learning curves across the 56 defined block
\ensuremath{\times} condition \ensuremath{\times} stimulus-type cells (10 bins \ensuremath{\times} 3 conditions \ensuremath{\times} 2 stimulus
types), while fitting a global lapse with a 31-point grid from 0 to 0.6.
The selected configuration was learning rate 0.30 with hidden noise 1.0,
at RMSE 0.077 \ensuremath{\pm} 0.001 over 15 seeds.

\subsubsection{Experiment 2 networks}\label{experiment-2-networks}

Exp. 2 networks were trained through both phases continuously. Each
network received its participant\textquotesingle s own pretraining
sequence (one-dimensional rule or XOR, according to that
participant\textquotesingle s assignment) and then that
participant\textquotesingle s R+E sequence. Weights carry over between
phases without reinitialisation, rescaling, or any change of learning
rate.

Every architectural and optimisation hyperparameter was inherited
unchanged from the Exp. 1 model (32 hidden units, learning rate 0.30,
initialisation scale 0.01, hidden noise 1.0). The only quantities
estimated from Exp. 2 are two lapse rates, 0.10 for pretraining and 0.24
for the R+E phase, reflecting the different difficulty of the two
phases. No parameter was fitted to the effect of interest.

\subsubsection{Measurement conventions for neural
networks}\label{measurement-conventions-for-neural-networks}

Learning curves are read from the training forward pass, with hidden
noise injected --- that is, networks are measured exactly as
participants are, trial by trial on the stimulus actually presented.
Test accuracy marginalises the hidden noise over 400 independent draws,
because the test phase asks what the network has learned rather than how
it behaves on a single noisy trial.

\subsubsection{Balanced washout
control}\label{balanced-washout-control}
 
To test whether test-phase differences between curricula reflect the
composition of the final training block, every Experiment 1 network was
retrained with an identical extra block of 48 trials appended after
block 5, regardless of condition: each of the six rule stimuli six
times and each exception six times, in an order shuffled independently
per network. All other settings were unchanged, and the test probe was
taken after the extra block.
 
\subsubsection{Frozen representation
control}\label{frozen-representation-control}
 
To test whether the pretraining $\times$ curriculum interaction requires
representational change, Experiment 2 was repeated with the
input-to-hidden weights and biases held at their initialisation and
only the readout trained. This is a fixed random-feature network, whose
participation ratio is constant. Because the hidden activations must be
on the scale of the injected noise for the readout to carry any signal,
this control uses an initialisation scale of 1.0 rather than 0.01
(placing it in the lazy regime); every other hyperparameter is
inherited unchanged.
 
Both controls were run over multiple simulation seeds (20 and 3
respectively), and accuracy was averaged across seeds before model
fitting.

\subsubsection{Representational analyses}\label{representational-analyses}

All representational measures were computed on the noise-free hidden
representations. I.e., the post-ReLU activations evoked by the eight
stimulus prototypes with no noise added.

\textbf{Dimensionality.} We summarised representational dimensionality
by the participation ratio, PR = (\ensuremath{\Sigma}\ensuremath{_{i}}\ensuremath{\lambda}\ensuremath{_{i}})\ensuremath{^{2}} / \ensuremath{\Sigma}\ensuremath{_{i}}\ensuremath{\lambda}\ensuremath{_{i}}\ensuremath{^{2}}, where \ensuremath{\lambda}\ensuremath{_{i}} are the
eigenvalues of the covariance of the eight mean-centred hidden
representations. PR ranges from 1 (all stimuli on a single axis) to 8
and is invariant to overall scaling of the representation.

\textbf{Task axes.} We additionally projected the hidden representations
onto two task-defined axes. The rule axis is the difference between the
mean representation of class-1 and class-0 stimuli over the six rule
stimuli. The exception axis is the direction separating the two
exception stimuli, orthogonalised against the rule axis and
renormalised. Both axes have a fixed sign in every network, so
projections can be averaged across networks. We report the separation
along each axis: for the rule axis, the class-1 minus class-0 mean over
rule stimuli and for the exception axis, the class-1 exception minus the
class-0 exception.

\subsubsection{Statistical analysis of network
results}\label{statistical-analysis-of-network-results}

Test accuracy was analysed with a linear mixed model fitted to one
observation per network per stimulus --- the network\textquotesingle s
probability of a correct response to that stimulus, marginalised over
the hidden noise and passed through the group lapse (Exp. 1: 1656
observations from 207 networks; Exp. 2: 8208 from 1026). This yields one
expected accuracy per stimulus. The models were accuracy
\textasciitilde{} curriculum * stimulus type + (1 + stimulus type
\textbar{} network) for Exp. 1 and, for Exp. 2, accuracy
\textasciitilde{} pretraining * curriculum * stimulus type + (1 +
stimulus type \textbar{} network), with stimulus type sum-coded.

Both models converge, though the estimated covariance is near-singular
(intercept SD \ensuremath{\approx} 0.003, slope SD \ensuremath{\approx} 0.007, against residual SDs of 0.078
and 0.087), because the networks are more homogeneous. Effects within
one stimulus type are reported as contrasts of these models, as the
corresponding human effects are; quantities from other analyses,
including learning curves and the representational measures, were tested
with Welch\textquotesingle s or paired t-tests. All estimates carry 95\%
confidence intervals.

\subsection{Data and code availability}\label{data-and-code-availability}

All research code and anonymized data is available on Github:
\url{https://github.com/licezhang/curriculum_unlearning}


\clearpage
\section{Supplementary information}\label{supplementary-information}

\setcounter{figure}{0}
\renewcommand{\thefigure}{S\arabic{figure}}

\begin{figure}[htbp]
\centering
\includegraphics[width=\linewidth]{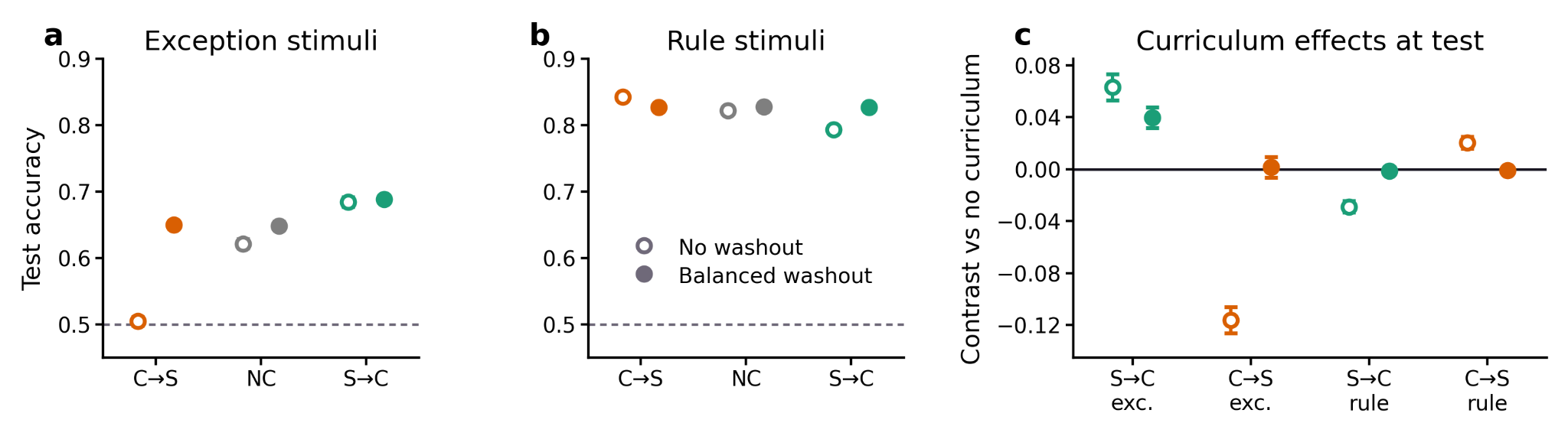}
\caption{\textbf{Balanced washout control.} We continued training
networks for an additional 48-trial block at the \emph{no-curriculum}
composition (36 rule, 12 exception) appended after block 5, so that all
three conditions end on the same local statistics; the test probe
follows that block. \textbf{a,} Test accuracy on exception stimuli by
curriculum, without (open) and with (filled) the washout block. Points
are means over 207 networks and error bars are 95\% confidence
intervals; accuracy is averaged over 20 simulation seeds before
plotting. \textbf{b,} The same for rule stimuli. \textbf{c,} Curriculum
contrasts against \emph{no curriculum}, estimated within stimulus type
from the same mixed model used for the network test-phase contrasts in
the main text (estimate and 95\% CI). The exception benefit of
\emph{simple-to-complex} is reduced but remains reliable after the
washout, whereas the \emph{complex-to-simple} exception cost and the
\emph{simple-to-complex} rule cost are both abolished, identifying them
as recency effects.}\label{fig:S1}
\end{figure}

\begin{figure}[htbp]
\centering
\includegraphics[width=\linewidth]{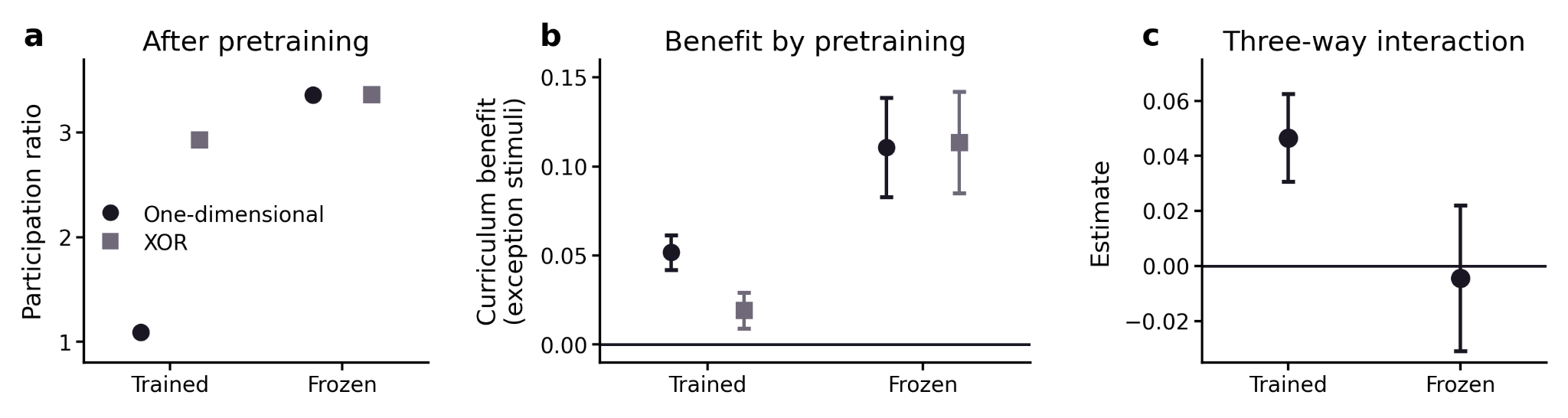}
\caption{\textbf{Frozen representation control.} Experiment 2 repeated
with the hidden layer frozen at its initialisation and only the readout
trained (a fixed random-feature network), alongside the
trained-representation model of Figure 5. \textbf{a,} Participation
ratio of the eight stimuli at the end of pretraining, by pretraining
task. One-dimensional pretraining compresses the trained network and
leaves the frozen network unchanged by construction. \textbf{b,}
Curriculum benefit on exception stimuli at test, within each
pretraining group (estimate and 95\% CI, from the same mixed model used
for the Experiment 2 test-phase contrasts in the main text). \textbf{c,}
The three-way pretraining \ensuremath{\times} curriculum
\ensuremath{\times} stimulus-type interaction for each network. The
frozen network shows a larger overall curriculum benefit than the
trained one, but it is the same after one-dimensional and XOR
pretraining and the interaction is absent. n = 1027 networks, three
simulation seeds.}\label{fig:S2}
\end{figure}

\end{document}